%% file: example_paper.tex
\documentclass{article}

\usepackage{microtype}
\usepackage{graphicx}
\usepackage{subfigure}
\usepackage{booktabs} 

\usepackage{hyperref}

\usepackage[accepted]{icml2024}

\usepackage{amsmath}
\usepackage{amssymb}
\usepackage{mathtools}
\usepackage{amsthm}

\usepackage{multirow}
\usepackage{enumitem}

\usepackage[capitalize,noabbrev]{cleveref}

\theoremstyle{plain}
\newtheorem{theorem}{Theorem}[section]

\theoremstyle{definition}

\theoremstyle{remark}

\usepackage[textsize=tiny]{todonotes}

\newcommand{\algname}{CTLPE} 

\newcommand{\algnamef}{NCDE-PE}

\icmltitlerunning{Continuous-Time Linear Positional Embedding for Irregular Time Series Forecasting}

\begin{document}

\twocolumn[
\icmltitle{Continuous-Time Linear Positional Embedding for Irregular Time Series Forecasting}



\icmlsetsymbol{equal}{*}

\begin{icmlauthorlist}
\icmlauthor{Byunghyun Kim}{sch}
\icmlauthor{Jae-Gil Lee}{sch}
\end{icmlauthorlist}

\icmlaffiliation{sch}{School of Computing, KAIST, Daejeon, Republic of Korea}

\icmlcorrespondingauthor{Jae-Gil Lee}{jaegil@kaist.ac.kr}

\icmlkeywords{Machine Learning, ICML}

\vskip 0.3in
]




\begin{figure*}[t!]
    \centering
    \includegraphics[trim=0cm 21.5cm 0cm 0cm, width=\textwidth, clip]{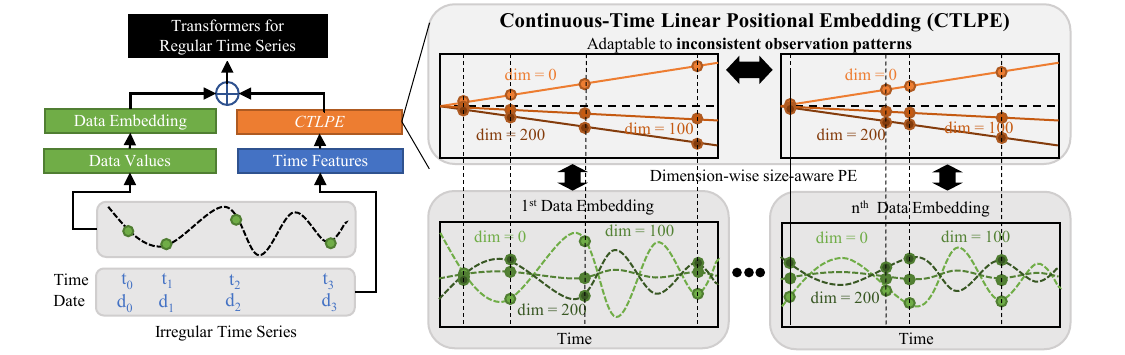}\\
    \caption{Overall framework of continuous-time learnable positional embedding (\algname{}).}
    \label{fig:overall}
    \vspace*{-0.1cm}
\end{figure*}

\begin{abstract}
Irregularly sampled time series forecasting, characterized by non-uniform intervals, is prevalent in practical applications. However, previous research have been focused on regular time series forecasting, typically relying on transformer architectures. To extend transformers to handle irregular time series, we tackle the positional embedding which represents the temporal information of the data. We propose \algname, a method learning a continuous linear function for encoding temporal information. The two challenges of irregular time series, inconsistent observation patterns and irregular time gaps, are solved by learning a continuous-time function and concise representation of position. Additionally, the linear continuous function is empirically shown superior to other continuous functions by learning a neural controlled differential equation-based positional embedding, and theoretically supported with properties of ideal positional embedding. \algname \ outperforms existing techniques across various irregularly-sampled time series datasets, showcasing its enhanced efficacy.
\end{abstract}

\input{1-Introduction}

\input{2-RelatedWork}
\input{3-Methodology}

\input{4-Evaluation}
\input{5-Conclusion}


\bibliography{example_paper}
\bibliographystyle{icml2024}


\input{6-Appendix}


\end{document}

%% file: 1-Introduction.tex
\section{Introduction}
\label{chap:introduction}

Irregularly sampled time series data, characterized by measurements taken at non-uniform intervals, is prevalent in practical scenarios with applications spanning medical monitoring \cite{silva2012predicting}, finance \cite{dal2015calibrating}, and environmental sensing \cite{menne2015united}. The accurate forecasting of outcomes in these diverse domains is crucial for avoidings risks and maximizing benefit. Previous research heavily focused on \emph{regular} time series forecasting, typically relying on transformer architectures. The self-attention mechanism of transformers demonstrate a remarkable ability to capture long-range dependencies and interactions, making them well-suited for time series analysis \cite{wen2022transformers}. However, these models for regular time series struggle with irregular time series due to their inability to accommodate a variety of inconsistent observation patterns and irregular time gaps.

The self-attention mechanism within the transformer model exhibits permutation invariance, signifying that self-attention inherently lacks knowledge of the sequential order in the data. Consequently, transformers rely on positional embeddings to integrate temporal information into time series analysis. Therefore, to adapt high-performing transformers designed for regular time series to irregular time series, modifying positional embedding is mandatory. By allowing the transformer to capture irregular temporal characteristics effectively through positional embedding, the self-attention of transformers can learn dependencies considering the temporal context of the time series.

To develop a positional embedding tailored for irregular time series, it becomes imperative to address two distinct challenges: inconsistent observation patterns and irregular time gaps. First, to provide positional embedding for any inconsistent observation patterns, positional embedding must be defined on continuous time. Otherwise, distinct observations share same parameters for representing positions which causes interference. Therefore, a learnable continuous function must be used as the positional embedding to handle any possible observation patterns. The second challenge is representing the irregular time gaps. Unlike regular time series or natural language which have equal temporal/semantic distance between neighboring tokens, the temporal context of irregular time series largely depend on the distance between observations. However, the self-attention of transformers do not have input space to directly learn from observed time gap.  Therefore, positional embedding for irregular time series should represent positional information concisely, which indirectly represent the irregular time gaps between observations. 

 Conventional positional embedding in regular time series forecasting consists of the widely adopted sinusoidal positional embedding\cite{vaswani2017attention} with uniform embedding\cite{zhou2021informer}  which embeds date information. The most popularly used positional embedding across multiple domains that use transformers is learnable positional embeddings, which introduce parameters that can adapt to data domain during the model training process. However, both of these conventional methods fail to solve the aformentioned challenges when applied to irregular time series data. First, both embedding consists of limited size of learnable parameters same as the input size. The limited number of parameters cannot cover all inconsistent observation patterns of irregular time series. Another reason is that exiting positional embeddings fail to capture irregular time gaps. This is because regular time series and other domains like natural language have consistent gaps between data points and therefore positional embedding is only designated to represent the order of the data.

To address the aforementioned challenges, we propose our method Continuous-Time Linear Positional Embedding (\algname), designed to tackle the intricacies associated with applying irregular time series data to transformers for regular time series. \algname \ is able to provide positional embedding value to any inconsistent observation patterns as it is defined on continuous-time. The linear form is derived from specific properties that facilitate the representation of position, monotonicity and translation invariance \cite{wang2020position}, effectively handling irregular time gaps. Furthermore, we design neural controlled differential equation-based positional embedding (\algnamef) to empirically prove that linear form is superior to any other continuous form of positional embeddings.

In summary, our primary contributions can be summarized as follows.
\begin{itemize}
\item Proposing \algname, a novel positional embedding method that bridges the gap between models for regular and irregular time series forecasting.
\item Demonstrating how \algname \ effectively represents positional information of irregular time series through learning a continuous function with theoretical basis of representing positions.
\item Using neural controlled differential equation-based positional embedding to show that linear form is superior to other continuous-time positional functions.
\item Outperforming existing positional embedding techniques when applied to forecasting tasks across various irregularly-sampled time series datasets, highlighting its enhanced efficacy.
\end{itemize}

%% file: 2-RelatedWork.tex
\section{Related Work}
\label{chap:related_work}

\subsection{Deep Learning Methods for Irregular Time Series Analysis}

To address the issues associated with irregular time series, deep learning methods have been employed to directly tackle tasks using irregular data. RNN-based methods have gained popularity in handling irregular time series  \cite{lipton2016directly} \cite{choi2016doctor}.  Among them, GRU-D \cite{che2018recurrent} has been notably successful in modeling irregular time series by incorporating decay into hidden states to capture irregularities. Nevertheless, a limitation arises as the hidden states undergo significant changes with each observation. In response to this constraint, neural differential equations have been introduced.

A neural differential equation (NDE) is a differential equation with a vector field that is parametrized by a neural network\cite{kidger2022neural}. The hidden state defined by a vector field is continuous, which allows the irregularly-sampled time series to be processed adequately. Neural ordinary differential equations (NODEs) \cite{chen2018neural} is the starting paper which define residual networks (ResNet) \cite{he2016identity} to continuous time, showing strength in modeling generative latent time series models. The proposed model is called Latent ODE which is based on the architecture of a variational autoencoder with a RNN Encoder and a ODE-based decoder. A following work by \cite{rubanova2019latent} further expand latent ODE's encoder to ODE-RNN, where the hidden states are defined continous respect to time. Neural controlled differential equations (NCDEs) \cite{kidger2020neural} reduce the ODE's limitation of high dependence on the initial condition by having a controlling path which considers all inputs. 

\subsection{Transformers in Time Series Forecasting}

Commencing with LogSparse Transformer \cite{li2019enhancing},  a range of transformer-based time series forecasting models has been introduced, including Informer \cite{zhou2021informer}, Autoformer \cite{wu2021autoformer}, Pyraformer \cite{liu2021pyraformer}, and Fedformer \cite{zhou2022fedformer}. LogSparse Transformer pioneered the application of transformer models to time series forecasting, leveraging their demonstrated excellence in tasks involving the capture of long-range dependencies. Notably, the memory cost is down from $\mathcal{O}(n^2)$ to $\mathcal{O}(n(\log n)^2)$  Informer proposed ProbSparse self-attention to alleviate time complexity down to $\mathcal{O}(n\log n)$. Autoformer employs decomposition methods to enhance performance, integrating an auto-correlation mechanism for identifying dependencies and facilitating information aggregation. Pyraformer further reduces self-attention time complexity to $\mathcal{O}(n)$ through the adoption of pyramidal attention module (PAM). Fedformer introduces frequency-enhanced blocks to capture the seasonality and trend of time series while maintaining a time complexity of $\mathcal{O}(n)$. 

Despite these advancements, transformers in time series forecasting face substantial criticism from LTSF-Linear \cite{zeng2023transformers}. Simple one-layer linear models from this research have been found to outperform all the aforementioned transformer models. However, transformer-based PatchTST \cite{nie2023a} employs patching and channel-independence to outperform linear models. Consequently, transformers in time series forecasting persist as state-of-the-art, necessitating further research to address irregularities.

\subsection{Positional Embedding in Transformers}

Positional embedding is a critical component in transformers as they analyze sequential data, and the self-attention mechanism is permutation invariant. In essence, without positional embedding, a transformer cannot discern the order of the provided data. Consequently, positional embeddings have undergone thorough development. In this section, we focus on some prominent ones in natural language processing and delve deeper into positional embeddings specifically designed for time series.

The sinusoidal positional embedding was initially introduced in the transformer paper \cite{vaswani2017attention}. The sinusoidal function carries various properties that is suitable for representing positions. Subsequently, the relative positional embedding \cite{shaw2018self} was introduced to represent relative gaps rather than absolute positions. While these embeddings are fixed and straightforward, they lack learnable parameters that can fit to data. Therefore, following the surge in transformer applications for pretrained language models like BERT \cite{devlin2018bert} and GPT \cite{radford2018improving}, these models began incorporating learnable positional embeddings. This simply adds a learnable linear layer to the input vector, which is then trained with the loss for each downstream task.

In the context of time series forecasting, specialized time-specific positional embeddings have been developed. The positional features initially devised for RNN-based time series forecasting \cite{salinas2020deepar} have been applied across all transformer-based time series forecasting works, from Logsparse Transformer to the recent PatchTST. Inputting year, month, day-of-the-week, hour-of-the-day, and minute-of-the-hour enables the learning of not only absolute time but also trends and seasonality from other temporal information.

%% file: 3-Methodology.tex
\section{Methodology}
\label{chap:method}

\subsection{Problem Definition: Irregular Time-Series Forecasting}

A multivariate irregular time series $\mathcal{T}$ can be defined as $\mathcal{T}=\{(t_1, x_1), (t_2, x_2) \cdots, (t_N, x_ N) \}$, where $x_i$ is a $l$-dimensional multivariate variable where each variable can be null. The timestamps $t_i \in \mathbb{R}$ are irregularly sampled such that $t_1 < \cdots< t_N$.

We further define the problem of \textit{irregular time-series forecasting}. Given an input lookback window, which forms a multivariate irregular time series $\mathcal{X}=\{(t_1, x_1), (t_2, x_2) \cdots, (t_N, x_ N) \}$, the goal is to predict subsequent time series of length $M$, $\mathcal{Y}=\{(t_{N+1}, x_{N+1}), (t_{N+2}, x_{N+2}) \cdots, (t_{N+M}, x_{N+M}) \}$.

\subsection{Overall framework of \algname \ with Transformers}

 The overall framework of how \algname \ operates within transformers for irregular time series forecasting is illustrated in Figure \ref{fig:overall}. \algname \ replaces the original positional embedding without making modifications to other components of the model. This adaptation enables the extension of transformer models designed for regular time series to effectively handle irregular time series. The time series for both past data and prediction is divided into two components: the data values and the time features. Data values represent the multivariate values of the time series, while time features solely encompass observation time-related information, including relative and absolute time and date.

The data values undergo embedding through the data embedding component of the model. Token embedding is a commonly employed technique in most existing transformer-based models for time series forecasting. This embedding utilizes convolutional layers to generate embeddings for each data point, allowing them to assimilate information from their adjacent data points. However, when dealing with irregular time series forecasting, the neighboring data points could be temporally distant. Therefore, we adapt the convolutional layer to consider only the information of the current data point, enabling other components of the model, including our \algname \ and attention, to learn from the embedding.

 Leveraging the time features of the time series, the positional embedding for irregular time series is generated using \algname. The detailed process of \algname \ is elaborated in \ref{sub:NCDE}. The irregularly spaced time features are learned within the vector space of neural differential equations, leading to the creation of the corresponding positional embedding. This positional embedding is then combined with the data embedding generated from token embedding and sent to the encoder of the transformer.

Although the encoder structure varies slightly among different papers on time series forecasting, they commonly incorporate the self-attention mechanism. The self-attention mechanism within the encoder is designed to understand the relationships between data points at various time intervals. The ability to grasp dependencies across different time points is vital for effective time series forecasting, as it is directly linked to the forecasting task.

 The decoder utilizes both the output from the encoder and the embedding of the time series for prediction, engaging in cross-attention. This process focuses on understanding the dependencies learned by the encoder and establishing connections with the time series designated for prediction. Subsequently, the prediction results are generated through a multi-layer perceptron. Mean Squared Error (MSE) loss is employed between the prediction results and the ground truth, and this loss is backpropagated through the decoder and all other components of the transformer, including our \algname.

\subsection{Continuous-Time Linear Positional Embedding (\algname)}
\label{sub:Linear}

While positional embeddings for regular time series only have to consider the order of the data, irregular times series needs to consider the irregular time gaps between observations. We show this in Appendix \ref{append:sin} that sinusoidal PE fails to represent irregular time gaps as the gap enlarges. Therefore, representing irregular time gap is mandatory for correctly encoding positional information of irregular time series.

A direct way to learn irregular time gap is to explicitly feed it into the deep learning model. However, transformers do not have architectural support to do this. Therefore, we indirectly learn time gap through providing precise position. How positional embedding can provide ideal positional information is studied in previous works, where monotonicity and translation invariant property stands as the ideal properties of positional embedding. \cite{wang2020position}. \\
\noindent \textbf{Definition 1. Monotonicity} : 
The distance between positional embeddings increases with larger time intervals.
\begin{align*}
\forall t_1, t_2, t_3 \in \mathbb{R}: ||t_1-t_2|| > ||t_1-t_3|| \\ 
\Leftrightarrow d(p(t_1), p(t_2)) > d(p(t_1), p(t_3)).
\end{align*}
This formally represents the requirement  for positional embeddings to preserve the sequential order of distances between positions. An embedding that adheres to monotonicity is commonly referred to as ordinal embedding. Monotonicity plays a crucial role in demonstrating that the embedding genuinely captures the positional information inherent in sequential data.

\noindent \textbf{Definition 2. Translation invariance} : The distance between positional embeddings are translation invariant.
\begin{align*}
\forall t_1, t_2, \dots, t_n, l \in \mathbb{R}: d(p(t_1), p(t_1+l)) \\ = d(p(t_2), p(t_2+l)) = \dots = d(p(t_n), p(t_n+l)).
\end{align*}
Positional embeddings that consider the time gaps between observations should adhere to the property of translation invariance. This is especially important in time series tasks, where a long time series is cut into windows. The translation invariant property insures that the positional information is provided correctly in any way the window is created. 

\begin{theorem}
\label{theorem:Linear}
{\sc (Linear Positional Embedding)} A positional embedding that satisfies monotonicity and translation invariance must be linear.
Formally,  \begin{equation}
\label{eq:ub_embedding_distance_main}
    p(x) = kx + b.
\end{equation}

\end{theorem}
\vspace*{-0.3cm}
\begin{proof} 
\begin{align*}
    &d(p(t_1), p(t_1+l)) \\ 
    &= d(p(t_2), p(t_2+l)) \\ 
    &= \dots = d(p(t_n), p(t_n+l)). & (by \ translation \ invariance) \\ 
    &\leftrightarrow ||p(t_i)-p(t_i+l)||^2 = K & (Eulidean \ dist, K\in \mathbb{R
^+}) \\
    &\leftrightarrow p(t_i+l) - p(t_i) = \sqrt{K} & (by \ monotonicity)
    \\
    &\leftrightarrow  p(t_i+2l) - p(t_i+l) = k & ( \sqrt{K} = k, t_i \rightarrow t_i+l)
     \\
    &p(t_i+2l)- p(t_i) = 2k
     \\
    &\leftrightarrow  p(t_i+xl)- p(t_i) = kx
     \\
    &\leftrightarrow  p(x)- p(0) = kx & (l = 1, t_i = 0) \\
    &\leftrightarrow  p(x) = kx + b   & (p(0) = b)
\end{align*}
\end{proof}

Linear positional embedding is created with two parameters, slope and bias as shown in Algorithm \ref{alg:CTLPE}. The slope and bias is learned for each data dimension of the data embedding. By learning these parameters through the loss function of the data, the positional embedding represents position without overwhelming the data embedding. We call this the \emph{size-aware} property of \algname \, where the positional embedding learns its size respect to the size of each data dimension in data embedding. 

\subsection{Neural Controlled Differential Equation-based Positional Embedding (\algnamef)}
\label{sub:NCDE}
To further prove that \algname \ is the best form of any learnable continuous time function, we propose neural-controlled-differential-equation-based positional embedding, \algnamef. By learning a vector field through neural differential equations, positional embedding can define any differentiable, therefore continuous function that is learnable from the data. By showing that \algnamef also learn a linear function, we prove that \algname effectively learns the best possible positional embedding for irregular time series.

\noindent \textbf{Definition of \algnamef}: $\mathcal{T}$ is cut into windows of $n$ timestamps, where a window is chosen as $w=\{(t_0, x_0), (t_1, x_1) \cdots, (t_{n-1}, x_{n-1}) \}$. Let the time feature $d_i$ corresponding to $t_i$, $\mathbb{d}=\{(t_0, d_0), (t_1, d_1) \cdots, \\  (t_{n-1}, d_ {n-1}) \}$. Then, $D:[t_0, t_{n-1}] \rightarrow \mathbb{R}^{m+1} $ is a natural cubic spline with knots at $t_0, \dots, t_{n-1}$. We further define two neural networks from the formation of Neural CDE \cite{kidger2020neural}. $f_{\theta_1}:\mathbb{R}^w \rightarrow  \mathbb{R}^{w \times m}$ a neural network with parameter $\theta_1$, and $\zeta_{\theta_2} : \mathbb{R}^{m} \rightarrow \mathbb{R}^w$ with $\theta_2$. Then, \algnamef \ is defined as the solution of CDE

\begin{equation}\label{eq_NCDEPE} 
p_t = p_{t_0} + \int_{t_0}^tf_{\theta_0}(p_s)dD_s, \quad p_{t_0} = \zeta_{\theta}(x_0, t_0).
\end{equation}

\hfill
\hfill

\noindent \textbf{Learning of \algnamef}: To demonstrate the learning of \algnamef, we first explain how neural CDE is trained. Similar to ODE, the adjoint backpropagation method can be applied to CDE, reducing the sequence length computation of neural differential equations' backpropagation to a single computation from end to the start of the sequence. This updates the parameters of the vector field $f_{\theta_1}$ and the initial state $\zeta_{\theta_2}$. Unlike previous works including the original neural CDE paper \cite{kidger2020neural} that use only the last state for classification,  \algnamef \ utilizes all hidden states  $p_{t_0}, p_{t_1}, \dots, p_{t_{n-1}} $ of neural CDE.This leads to larger computation for each positional embedding during backpropagation. However, this computation is crucial for learning positional embedding and enables faster training of \algnamef, incorporating techniques for accelerated training as explained in the following section.

Most importantly, the learning of \algnamef \ can be conducted on any irregularly spaced time series window. The learning domain of $p_{t}$ is defined on any time sequence, as a vector space is learned instead of a fixed neural network. Consequently, \algnamef \ learns a suitable positional embedding for irregular time series. Furthermore, the continuous vector field of neural CDE naturally predicts positional embedding for unseen time stamps in training time.  \\

\input{algorithms/TSPE}
\input{algorithms/CTLPE}
\input{algorithms/NCDEPE}

\noindent \textbf{Managing computational cost}: Despite the advantages that differential equations offer for irregular time series forecasting, the recurrent calculations involved in neural differential equations result in extended training times, approximately $\mathcal{O}(n)$. Additionally, the Runge-Kutta method, employed for integration approximation, may take considerable time depending on inputs. In contrast, original fixed embeddings or learnable embeddings lack recurrent calculations between embeddings, enabling parallel computation with a time complexity of $\mathcal{O}(1)$. To address this challenge, we employ two techniques.

The initial technique involves storing \algnamef \ as a table. The embedding values at the desired time can be precomputed and stored in a table. Consequently during test time, the positional embedding is no longer being learned. This approach ensures that during inference, the positional embedding does not necessitate recurrent computation but merely involves reading values from the positional embedding table. This enables parallel computation of self-attention, reducing the computation cost to $\mathcal{O}(1)$. 

To further decrease computational time, we leverage the fact that \algnamef \ has significantly fewer learnable parameters in comparison to the rest of the transformer architecture. This implies that the positional embedding can be trained much more quickly than the parameters involved in self-attention. Additionally, we only train \algnamef \ for a single epoch. Subsequently, the embedding is saved as a table, enabling parallel computation for the remaining epochs during training, also at $\mathcal{O}(1)$ complexity. \\

\noindent \textbf{Managing memory cost}: The learning through backpropagation of neural differential equations is more memory-efficient compared to other neural networks. General neural networks, when trained over the sequence length $t$ and memory size of the vector field of neural differential equations $M$, incur a memory cost of $\mathcal{O}(tM)$. However, leveraging the adjoint method from neural ordinary differential equations \cite{chen2018neural}, the backpropagation of ODE can be computed from the end to the start of the sequence in a single pass, reducing the memory cost to $\mathcal{O}(M)$. For Neural CDE, this cost is slightly increased to keep the time series data, resulting in a memory cost of $\mathcal{O}(M+t)$. \\

%% file: algorithms/TSPE.tex
\begin{algorithm}[t]
\caption{Overall procedure for transformer-based irregular time series forecasting with \algname}
\label{alg:TSPE}
    \begin{algorithmic}[1]
        \REQUIRE past time window $w_p$, predicting time window $w_f$, past time feature $d_p$, predicting time feature $d_f$, encoder layer depth N, decoder layer depth M
        \STATE{$\mathcal{X}_{en, val} = $ TokenEmbed$_{en}(w_f)$ \COMMENT{Value embedding for encoder} }
        \STATE{$\mathcal{X}_{de, val} = $ TokenEmbed$_{de}(w_p)$ \COMMENT{Value embedding for decoder} }
        \STATE{$\mathcal{X}_{en, pos} = \algname_{en}(d_p)$} \COMMENT{Positional embedding for encoder} 
        \STATE{$\mathcal{X}_{de, pos} = \algname_{de}(d_f)$} \COMMENT{Positional embedding for decoder} 
        \STATE{$\mathcal{X}_{en}^0 = \mathcal{X}_{en, val} + \mathcal{X}_{en, pos}$}
        \FOR{$l=1$ {\bf to} $l=N$} 
            \STATE {$\mathcal{S}_{en}^{l, 1} =$ Self-Attention$(\mathcal{X}_{en}^{l-1}) + \mathcal{X}_{en}^{l-1}$ } \COMMENT{Encoder self-attention} 
            \STATE {$\mathcal{S}_{en}^{l, 2} =$ FeedForward$(\mathcal{S}_{en}^{l, 1}) + \mathcal{S}_{en}^{l, 1} $ }
        \ENDFOR
        \STATE{$\mathcal{X}_{de}^0 = \mathcal{X}_{de, val} + \mathcal{X}_{de, pos}$}
        \FOR{$l=1$ {\bf to} $l=M$} 
            \STATE {$\mathcal{S}_{de}^{l, 1} =$ Self-Attention$(\mathcal{X}_{de}^{l-1}) + \mathcal{X}_{de}^{l-1}$ } \COMMENT{Decoder self-attention} 
            \STATE {$\mathcal{S}_{de}^{l, 2} =$ Cross-Attention$(\mathcal{X}_{de}^{l-1}, \mathcal{X}_{en}^N) + \mathcal{X}_{de}^{l-1}$ } \COMMENT{Decoder cross-attention} 
            \STATE {$\mathcal{S}_{de}^{l, 3} =$ FeedForward$(\mathcal{S}_{de}^{l, 1}) + \mathcal{S}_{de}^{l, 1} $ }
            \STATE {$\mathcal{X}_{de}^l = \mathcal{S}_{de}^{l, 3}$}
        \ENDFOR
        \STATE {$\mathcal{X}_{pred} =$ MLP($\mathcal{X}_{de}^M$)}
        \STATE \textbf{Return} $\mathcal{X}_{pred}$
    \end{algorithmic}
\end{algorithm}

%% file: algorithms/CTLPE.tex
\begin{algorithm}[t]
\caption{\algname{} algorithm}
\label{alg:CTLPE}
    \begin{algorithmic}[1]
        \REQUIRE time $\textbf{t}$ with $n$ timestamps, learnable parameters slope \textbf{a} and bias \textbf{b}
        \STATE $\mathcal{X}_{pos} = \textbf{a}*\textbf{t} + \textbf{b}$
        \STATE \textbf{Return} $\mathcal{X}_{pos}$
    \end{algorithmic}
\end{algorithm}

%% file: algorithms/NCDEPE.tex
\begin{algorithm}[t]
\caption{\algnamef{} algorithm}
\label{alg:NCDEPE}
    \begin{algorithmic}[1]
        \REQUIRE time feature $d$ with $n$ timestamps, 
        \STATE $D$ = NaturalCubicSpline($d$)
        \STATE $p_{t_1} = \zeta_{\theta}(x_0, t_0)$
        \FOR{$i=2$ {\bf to} $i=n$} 
            \STATE {$p_{t_i} = p_{t_{i-1}} + \int_{t_{i-1}}^{t_i}f_{\theta_0}(p_s)dD_s$} \COMMENT {Integral calculated with fourth-order Runge-Kutta} 
        \ENDFOR
        \STATE {$\mathcal{X}_{pos} = \{p_{t_1}, p_{t_2}, \dots, p_{t_n}\}$}
        \STATE \textbf{Return} $\mathcal{X}_{pos}$
    \end{algorithmic}
\end{algorithm}

%% file: 4-Evaluation.tex
\input{tables/tab_total}
\section{Evaluation}
\label{chap:eval}
\subsection{Experiment Setting}

\subsubsection{Data Description} 

\textbf{ETT} \cite{zhou2021informer} refers to Electricity Transformer Temperature data, which holds significant importance in electric power deployment. This dataset constitutes regular time series data spanning 2 years, encompassing two counties named ETT$\rm h_1$ and ETT$\rm h_2$ with hourly resolution, and ETT$\rm m_1$ with 15-minute resolution. The dataset offers both multivariate (ETT$\rm h_1$M, ETT$\rm h_2$M) and univariate (ETT$\rm h_1$S, ETT$\rm h_2$S) types. As in previous studies, the train, validation, and test sets are divided into 12/4/4 months each.

\textbf{Weather} dataset comprises hourly climate data collected over 4 years in the U.S. from the National Oceanic and Atmospheric Administration (NOAA). This dataset includes 11 features with a single target value, "wet bulb." Similar to previous work, the train, validation, and test sets are divided into 28/10/10 months each.

\textbf{ECL} \cite{misc_electricityloaddiagrams20112014_321} is Electricity Consuming Load data obtained from the UCI Machine Learning Repository. It features 15-minute-sampled electricity consumption data converted to hourly data from 321 users. Following the convention in earlier studies, the train, validation, and test sets are divided into 15/3/4 months each.

To introduce irregularity into the time series data, we randomly drop data points from each dataset, resulting in irregular time series data with irregularities of 20\%, 40\%, and 60\%. The purpose of creating irregular data from regular data is to enable the use of other positional embedding methods, such as sinusoidal and time features.

\subsubsection{Evaluation Environment}
We use Informer \cite{zhou2021informer} and FEDformer \cite{zhou2022fedformer} as base model for applying our \ \algname. RevIN (Reversible Instance Normalization) \cite{kim2021reversible} is used to solve distribution shift problem in Informer. MSE (mean squared error) and MAE(mean absolute error) are used as evaluation metrics. The experiments are ran three times and their average and standard deviation are provided.

\subsubsection{Baselines}
\begin{itemize}
    \item \emph{\textbf{Uniform representation embedding}} introduced by Informer \cite{zhou2021informer}. This is primarily designed for regular time series forecasting, which utilizes not only time but also date features for creating the positional embedding.
    \item \emph{\textbf{Sinusoidal positional embedding}} from original transformer paper \cite{vaswani2017attention} is also used due to their demonstrated effectiveness across various tasks. An extended version, \emph{Irr-sinusoidal} is a sinusoidal positional embedding method extended for irregular time series by indexing the sinusoidal function with observation time. 
    \item  \emph{\textbf{Time feature embedding}} \cite{shukla2021multi} is multidimensional time features as input and employing learnable embeddings to enhance temporal information. 
    \item  \emph{\textbf{Simple learnable embedding}} Utilizing simple learnable embedding as same as the size of the data embedding is a widely used techinque in natural language processing (BERT \cite{devlin2018bert}) and large language models (GPT \cite{brown2020language}). We create two versions of this method. \emph {Simple} is where the learnable parameters are dedicated for each possible observation time. Therefore, only few parameters that correspond to the observation time are trained during the learning process. \emph {Simple overlap} on the other hand only considers the order of the time series, where learned positional embedding parameters correspond to the order in each time window.
\end{itemize}

\subsection{Experimental Results}
\label{4.2}
Table \ref{tab:total} and Table \ref{tab:FED} each show irregular multivariate time-series forecasting applied on Informer and FEDformer. Our \algname \ outperforms all other baselines in various missing rates for most of the cases on both architectures. Also, multi-dimensional time feature based method perform worst for most of the cases.     

Table \ref{tab:ETTH1M}, Table \ref{tab:ETTH2M}, and Table \ref{tab:WTHM} present the irregular multivariate time-series forecasting results in Informer across three datasets, considering various prediction lengths. The prediction length refer to 1 day, 2 day, 1 week, and 2 weeks each, following the convention of regular time series forecasting works. \algname \ functions effectively regardless of the prediction length of the forecasting tasks.

\input{tables/tab_abl_bias}

\subsection{Qualitative Analysis}

Figure \ref{fig:PEvisual} compares the sinusoidal positional embedding and \algname. Sinusoidal positional embedding provides different frequencies of sin graph for each data dimension. \algname \ instead learns a linear function with different slope value for each data dimension. By learning simple linear function instead of complex forms, the positional embedding is able to provide more direct positional information in irregular circumstances.

Experimentally, we observe that \algname \ learns the positional embedding in the below mathematical form for irregular time series forecasting. 
\begin{equation}\label{eq_NCDEPE_ex} 
p_i = (k_1t_i, k_2t_i, ... , k_mt_i) \\
    = \textbf{k}t_i.
\end{equation}
For m-dimensional positional embedding, the \algname \ learns the slope for each dimension and provides value that is proportional to time. Using this mathematical form, we prove that \algname \ satisfies variety of ideal properties of positional embedding in Appendix.

%% file: tables/tab_total.tex
\begin{table*}[ht]
    \caption{Forecasting results based on Informer + RevIN averaged on all datasets .}
    \centering
    \vspace*{0.2cm}
    \resizebox{\textwidth}{!}{%
    \begin{tabular}{@{} c|c | cc | cc | cc | cc | cc | cc | cc | cc @{}}
    \toprule
    \multicolumn{2}{c|}{\textbf{Methods}} &
    \multicolumn{2}{c|}{\textbf{Uniform}} &
    \multicolumn{2}{c|}{\textbf{Sinusoidal}} &
    \multicolumn{2}{c|}{\textbf{Irr-sinusoidal}} &
    \multicolumn{2}{c|}{\textbf{Time Feature}} &
    \multicolumn{2}{c|}{\textbf{Simple}} &
    \multicolumn{2}{c|}{\textbf{Simple overlap}} &
    \multicolumn{2}{c|}{\textbf{\algnamef{}}}  &
    \multicolumn{2}{c|}{\textbf{Linear}} 
     \\\cline{1-18}
    Prediction Length & Missing Rate(\%) & MSE & MAE & MSE & MAE& MSE & MAE& MSE & MAE& MSE & MAE & MSE & MAE & MSE & MAE & MSE & MAE   \\

\midrule
\multirow{8}{*}{24}
& \multirow{2}{*}{0} 
& 0.490& 0.471& 0.466& 0.461& 0.483& 0.469& 0.596& 0.519& 0.355& 0.392& 0.363& 0.397& 0.358& 0.391& \textbf{0.312}& \textbf{0.360}\\
&& $(\pm0.209)$& $(\pm0.124)$& $(\pm0.190)$& $(\pm0.115)$& $(\pm0.201)$& $(\pm0.121)$& $(\pm0.243)$& $(\pm0.123)$& $(\pm0.089)$& $(\pm0.047)$& $(\pm0.100)$& $(\pm0.052)$& $(\pm0.090)$& $(\pm0.057)$& $(\pm0.089)$& $(\pm0.055)$\\& \multirow{2}{*}{20} 
& 0.570& 0.509& 0.555& 0.505& 0.579& 0.519& 0.602& 0.524& 0.466& 0.456& 0.464& 0.452& 0.439& 0.436& \textbf{0.427}& \textbf{0.428}\\
&& $(\pm0.222)$& $(\pm0.119)$& $(\pm0.197)$& $(\pm0.109)$& $(\pm0.202)$& $(\pm0.114)$& $(\pm0.216)$& $(\pm0.114)$& $(\pm0.175)$& $(\pm0.084)$& $(\pm0.190)$& $(\pm0.094)$& $(\pm0.170)$& $(\pm0.088)$& $(\pm0.176)$& $(\pm0.093)$\\& \multirow{2}{*}{40} 
& 0.639& 0.549& 0.635& 0.548& 0.638& 0.552& 0.632& 0.548& 0.553& 0.509& 0.540& 0.495& 0.513& 0.478& \textbf{0.489}& \textbf{0.470}\\
&& $(\pm0.232)$& $(\pm0.120)$& $(\pm0.239)$& $(\pm0.126)$& $(\pm0.227)$& $(\pm0.122)$& $(\pm0.212)$& $(\pm0.114)$& $(\pm0.223)$& $(\pm0.108)$& $(\pm0.240)$& $(\pm0.110)$& $(\pm0.205)$& $(\pm0.102)$& $(\pm0.203)$& $(\pm0.104)$\\& \multirow{2}{*}{60} 
& 0.666& 0.570& 0.662& 0.569& 0.676& 0.577& 0.678& 0.580& 0.612& 0.544& 0.616& 0.546& 0.550& 0.504& \textbf{0.538}& \textbf{0.498}\\
&& $(\pm0.188)$& $(\pm0.104)$& $(\pm0.217)$& $(\pm0.116)$& $(\pm0.225)$& $(\pm0.121)$& $(\pm0.192)$& $(\pm0.106)$& $(\pm0.185)$& $(\pm0.093)$& $(\pm0.227)$& $(\pm0.108)$& $(\pm0.162)$& $(\pm0.079)$& $(\pm0.161)$& $(\pm0.081)$\\
\midrule
\multirow{8}{*}{48}
& \multirow{2}{*}{0} 
& 0.543& 0.507& 0.545& 0.503& 0.538& 0.500& 0.649& 0.555& 0.480& 0.460& 0.477& 0.466& 0.416& 0.429& \textbf{0.413}& \textbf{0.427}\\
&& $(\pm0.147)$& $(\pm0.085)$& $(\pm0.161)$& $(\pm0.089)$& $(\pm0.154)$& $(\pm0.088)$& $(\pm0.221)$& $(\pm0.112)$& $(\pm0.135)$& $(\pm0.064)$& $(\pm0.131)$& $(\pm0.067)$& $(\pm0.069)$& $(\pm0.037)$& $(\pm0.088)$& $(\pm0.049)$\\& \multirow{2}{*}{20} 
& 0.643& 0.550& 0.665& 0.559& 0.659& 0.556& 0.653& 0.562& 0.617& 0.532& 0.583& 0.520& 0.531& 0.488& \textbf{0.525}& \textbf{0.487}\\
&& $(\pm0.205)$& $(\pm0.101)$& $(\pm0.183)$& $(\pm0.083)$& $(\pm0.200)$& $(\pm0.092)$& $(\pm0.161)$& $(\pm0.083)$& $(\pm0.235)$& $(\pm0.100)$& $(\pm0.171)$& $(\pm0.075)$& $(\pm0.169)$& $(\pm0.079)$& $(\pm0.165)$& $(\pm0.078)$\\& \multirow{2}{*}{40} 
& 0.705& 0.590& 0.694& 0.585& 0.710& 0.595& 0.742& 0.607& 0.641& 0.561& 0.688& 0.576& 0.600& 0.530& \textbf{0.560}& \textbf{0.506}\\
&& $(\pm0.177)$& $(\pm0.087)$& $(\pm0.170)$& $(\pm0.090)$& $(\pm0.180)$& $(\pm0.091)$& $(\pm0.165)$& $(\pm0.082)$& $(\pm0.183)$& $(\pm0.090)$& $(\pm0.265)$& $(\pm0.120)$& $(\pm0.146)$& $(\pm0.068)$& $(\pm0.123)$& $(\pm0.052)$\\& \multirow{2}{*}{60} 
& 0.698& 0.596& 0.732& 0.609& 0.702& 0.595& 0.711& 0.600& 0.686& 0.590& 0.640& 0.568& 0.589& 0.535& \textbf{0.555}& \textbf{0.512}\\
&& $(\pm0.163)$& $(\pm0.088)$& $(\pm0.193)$& $(\pm0.101)$& $(\pm0.199)$& $(\pm0.103)$& $(\pm0.170)$& $(\pm0.092)$& $(\pm0.215)$& $(\pm0.104)$& $(\pm0.163)$& $(\pm0.083)$& $(\pm0.120)$& $(\pm0.063)$& $(\pm0.116)$& $(\pm0.051)$\\
\midrule
\multirow{8}{*}{168}
& \multirow{2}{*}{0} 
& 0.703& 0.597& 0.713& 0.601& 0.730& 0.604& 0.764& 0.628& 0.701& 0.598& 0.720& 0.603& 0.695& 0.588& \textbf{0.628}& \textbf{0.559}\\
&& $(\pm0.092)$& $(\pm0.050)$& $(\pm0.076)$& $(\pm0.050)$& $(\pm0.122)$& $(\pm0.068)$& $(\pm0.150)$& $(\pm0.078)$& $(\pm0.050)$& $(\pm0.023)$& $(\pm0.071)$& $(\pm0.018)$& $(\pm0.175)$& $(\pm0.080)$& $(\pm0.144)$& $(\pm0.075)$\\& \multirow{2}{*}{20} 
& 0.769& 0.637& 0.819& 0.658& 0.779& 0.637& 0.800& 0.651& 0.762& 0.634& 0.774& 0.639& 0.678& 0.588& \textbf{0.639}& \textbf{0.566}\\
&& $(\pm0.229)$& $(\pm0.110)$& $(\pm0.244)$& $(\pm0.117)$& $(\pm0.246)$& $(\pm0.118)$& $(\pm0.233)$& $(\pm0.115)$& $(\pm0.246)$& $(\pm0.121)$& $(\pm0.229)$& $(\pm0.114)$& $(\pm0.212)$& $(\pm0.104)$& $(\pm0.100)$& $(\pm0.048)$\\& \multirow{2}{*}{40} 
& 0.821& 0.665& 0.844& 0.672& 0.843& 0.672& 0.783& 0.647& 0.785& 0.651& 0.850& 0.679& 0.655& 0.584& \textbf{0.617}& \textbf{0.560}\\
&& $(\pm0.246)$& $(\pm0.122)$& $(\pm0.260)$& $(\pm0.130)$& $(\pm0.284)$& $(\pm0.135)$& $(\pm0.234)$& $(\pm0.116)$& $(\pm0.294)$& $(\pm0.141)$& $(\pm0.306)$& $(\pm0.143)$& $(\pm0.159)$& $(\pm0.081)$& $(\pm0.093)$& $(\pm0.039)$\\& \multirow{2}{*}{60} 
& 0.812& 0.674& 0.795& 0.663& 0.800& 0.666& 0.798& 0.663& 0.766& 0.650& 0.754& 0.645& 0.693& 0.604& \textbf{0.625}& \textbf{0.572}\\
&& $(\pm0.149)$& $(\pm0.070)$& $(\pm0.180)$& $(\pm0.092)$& $(\pm0.179)$& $(\pm0.093)$& $(\pm0.183)$& $(\pm0.092)$& $(\pm0.223)$& $(\pm0.109)$& $(\pm0.220)$& $(\pm0.109)$& $(\pm0.147)$& $(\pm0.066)$& $(\pm0.068)$& $(\pm0.036)$\\
\midrule
\multirow{8}{*}{336}
& \multirow{2}{*}{0} 
& 0.791& 0.643& 0.794& 0.645& 0.841& 0.663& 0.814& 0.660& 0.833& 0.662& 0.877& 0.683& 0.687& 0.599& \textbf{0.655}& \textbf{0.581}\\
&& $(\pm0.102)$& $(\pm0.057)$& $(\pm0.080)$& $(\pm0.051)$& $(\pm0.147)$& $(\pm0.074)$& $(\pm0.172)$& $(\pm0.088)$& $(\pm0.210)$& $(\pm0.090)$& $(\pm0.229)$& $(\pm0.104)$& $(\pm0.138)$& $(\pm0.074)$& $(\pm0.165)$& $(\pm0.089)$\\& \multirow{2}{*}{20} 
& 0.826& 0.671& 0.856& 0.687& 0.868& 0.692& 0.820& 0.668& 0.881& 0.687& 0.848& 0.677& 0.696& 0.600& \textbf{0.638}& \textbf{0.572}\\
&& $(\pm0.194)$& $(\pm0.103)$& $(\pm0.219)$& $(\pm0.112)$& $(\pm0.214)$& $(\pm0.112)$& $(\pm0.196)$& $(\pm0.105)$& $(\pm0.225)$& $(\pm0.109)$& $(\pm0.213)$& $(\pm0.109)$& $(\pm0.175)$& $(\pm0.083)$& $(\pm0.099)$& $(\pm0.055)$\\& \multirow{2}{*}{40} 
& 0.893& 0.714& 0.950& 0.736& 0.931& 0.733& 0.922& 0.724& 0.874& 0.698& 0.910& 0.717& 0.703& 0.611& \textbf{0.666}& \textbf{0.594}\\
&& $(\pm0.265)$& $(\pm0.138)$& $(\pm0.249)$& $(\pm0.130)$& $(\pm0.252)$& $(\pm0.129)$& $(\pm0.230)$& $(\pm0.116)$& $(\pm0.207)$& $(\pm0.095)$& $(\pm0.279)$& $(\pm0.131)$& $(\pm0.168)$& $(\pm0.097)$& $(\pm0.099)$& $(\pm0.060)$\\& \multirow{2}{*}{60} 
& 0.988& 0.761& 0.994& 0.768& 0.996& 0.768& 1.005& 0.770& \textbf{0.862}& 0.700& 0.880& 0.706& 0.864& \textbf{0.691}& 1.145& 0.798\\
&& $(\pm0.257)$& $(\pm0.134)$& $(\pm0.225)$& $(\pm0.119)$& $(\pm0.218)$& $(\pm0.115)$& $(\pm0.235)$& $(\pm0.125)$& $(\pm0.154)$& $(\pm0.069)$& $(\pm0.199)$& $(\pm0.091)$& $(\pm0.159)$& $(\pm0.087)$& $(\pm0.574)$& $(\pm0.231)$\\

    \bottomrule
    \end{tabular}%
    }
    \label{tab:total}
\end{table*}

%% file: tables/tab_abl_bias.tex
\begin{table}[ht]
    \centering
    \caption{ Ablation study on bias of \algname{} with forecasting results based on Informer + RevIN averaged on all datasets.}
    \vspace*{0.2cm}
    \resizebox{\textwidth/2}{!}{%
    \begin{tabular}{@{} c|c | cc | cc  @{}}
    \toprule
    \multicolumn{2}{c|}{\textbf{Methods}} &
    \multicolumn{2}{c|}
    {\textbf{\algname{}}}  &
    \multicolumn{2}{c}{\textbf{\algname{}} without bias} 
     \\\cline{1-6}
    Prediction Length & Missing Rate(\%) & MSE & MAE & MSE & MAE \\

\midrule
\multirow{8}{*}{24}
& \multirow{2}{*}{0} 
& \textbf{0.312}& \textbf{0.360}& 0.315& 0.362\\
&& $(\pm0.089)$& $(\pm0.055)$& $(\pm0.093)$& $(\pm0.059)$\\& \multirow{2}{*}{20} 
& 0.427& 0.428& \textbf{0.426}& \textbf{0.427}\\
&& $(\pm0.176)$& $(\pm0.093)$& $(\pm0.176)$& $(\pm0.092)$\\& \multirow{2}{*}{40} 
& \textbf{0.489}& \textbf{0.470}& 0.497& 0.474\\
&& $(\pm0.203)$& $(\pm0.104)$& $(\pm0.205)$& $(\pm0.106)$\\& \multirow{2}{*}{60} 
& 0.538& 0.498& \textbf{0.531}& \textbf{0.490}\\
&& $(\pm0.161)$& $(\pm0.081)$& $(\pm0.162)$& $(\pm0.077)$\\
\midrule
\multirow{8}{*}{48}
& \multirow{2}{*}{0} 
& \textbf{0.413}& \textbf{0.427}& 0.433& 0.437\\
&& $(\pm0.088)$& $(\pm0.049)$& $(\pm0.101)$& $(\pm0.055)$\\& \multirow{2}{*}{20} 
& 0.525& \textbf{0.487}& \textbf{0.523}& 0.488\\
&& $(\pm0.165)$& $(\pm0.078)$& $(\pm0.168)$& $(\pm0.081)$\\& \multirow{2}{*}{40} 
& 0.560& 0.506& \textbf{0.543}& \textbf{0.498}\\
&& $(\pm0.123)$& $(\pm0.052)$& $(\pm0.144)$& $(\pm0.064)$\\& \multirow{2}{*}{60} 
& \textbf{0.555}& \textbf{0.512}& 0.563& 0.516\\
&& $(\pm0.116)$& $(\pm0.051)$& $(\pm0.114)$& $(\pm0.050)$\\
\midrule
\multirow{8}{*}{168}
& \multirow{2}{*}{0} 
& \textbf{0.628}& \textbf{0.559}& 0.642& 0.571\\
&& $(\pm0.144)$& $(\pm0.075)$& $(\pm0.138)$& $(\pm0.067)$\\& \multirow{2}{*}{20} 
& 0.639& 0.566& \textbf{0.625}& \textbf{0.564}\\
&& $(\pm0.100)$& $(\pm0.048)$& $(\pm0.102)$& $(\pm0.043)$\\& \multirow{2}{*}{40} 
& \textbf{0.617}& 0.560& 0.620& \textbf{0.560}\\
&& $(\pm0.093)$& $(\pm0.039)$& $(\pm0.090)$& $(\pm0.038)$\\& \multirow{2}{*}{60} 
& 0.625& 0.572& \textbf{0.622}& \textbf{0.569}\\
&& $(\pm0.068)$& $(\pm0.036)$& $(\pm0.067)$& $(\pm0.034)$\\
\midrule
\multirow{8}{*}{336}
& \multirow{2}{*}{0} 
& \textbf{0.655}& \textbf{0.581}& 0.657& 0.586\\
&& $(\pm0.165)$& $(\pm0.089)$& $(\pm0.184)$& $(\pm0.106)$\\& \multirow{2}{*}{20} 
& 0.638& 0.572& \textbf{0.620}& \textbf{0.565}\\
&& $(\pm0.099)$& $(\pm0.055)$& $(\pm0.123)$& $(\pm0.065)$\\& \multirow{2}{*}{40} 
& 0.666& 0.594& \textbf{0.639}& \textbf{0.583}\\
&& $(\pm0.099)$& $(\pm0.060)$& $(\pm0.078)$& $(\pm0.040)$\\& \multirow{2}{*}{60} 
& 1.145& 0.798& \textbf{0.794}& \textbf{0.664}\\
&& $(\pm0.574)$& $(\pm0.231)$& $(\pm0.169)$& $(\pm0.083)$\\

    \bottomrule
    \end{tabular}%
    }
    \label{tab:ablation}
\end{table}

%% file: 5-Conclusion.tex
\section{Conclusion}
\label{chap:conclusion}

In this paper, we present \algname, a linear learnable embedding for irregular time series forecasting using transformers. By only modifying the positional embedding to our method, the architectures designed for regular time series were adapted to handle irregular time series. This positional embedding method outperformed any other fixed or learnable embeddings from both time series analysis and natural language processing.

For future works, we plan to work on transformer-based time-series forecasting models such as PatchTST \cite{nie2023a}. While PatchTST have shown the state-of-the-art in current time series forecasting models, the patching process adds additional dimension blocking a smooth adapation of \algname \ to PatchTST. Furthermore, we will apply \algname \ to other downstream tasks of time series. The adaptability to inconsistent observation time and the ability to represent irregular time gaps not only benefit forecasting tasks, but other time series tasks like classification or anomaly detection.

\section*{Acknowledgements}
This work was supported by Institute of Information \& Communications Technology Planning \& Evaluation\,(IITP) grant funded by the Korea government\,(MSIT) (No. 2020-0-00862, DB4DL: High-Usability and Performance In-Memory Distributed DBMS for Deep Learning).

%% file: 6-Appendix.tex
\newpage
\appendix
\onecolumn
\section{Appendix}

\subsection{Experimetal Results}

\input{tables/tab_ETTh1M}
\input{tables/tab_ETTh2M}
\input{tables/tab_WTHM}
\input{tables/tab_fed_ETTh1M}

\subsection{Limitations of sinusoidal positional embeddings for irregular time series.}
\label{append:sin}
\begin{figure*}[t!]
    \centering
    \includegraphics[trim=0cm 22.5cm 0cm 0cm, width=\textwidth, clip]{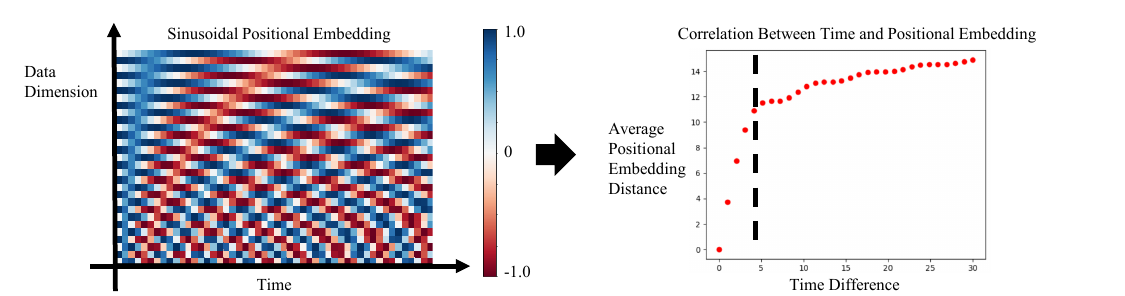}\\
    
    \caption{Sinusoidal positional embedding's correlation between time difference and average positional embedding distance.}
    \label{fig:pevis}
    \vspace*{-0.1cm}
\end{figure*}
One critical reason why existing positional embedding methods fail to support irregular time series was due to their inability to represent irregular time gaps. To support this statement, we show this problem with sinusoidal positional embedding since other methods are learnable methods having inconsistent representations.

\subsection{Properties of Ideal Positional Embedding}
While the definition of ideal positional embedding is controversial and there is no generally accepted concept, there are variety of properties that have been proposed. Therefore, we organize the properties from previous papers and also specific for irregular time series forecasting. 

For time $t \in \mathbb{R}$, we define the corresponding positional embedding as $p(t)$. Also, we define a distance measure $d$, where smaller the $d(p(t_1), p(t_2))$ means that the embeddings are more similar. The upcoming first three properties from \cite{wang2020position} reveal how positional embeddings represent position well. 

\noindent \textbf{Property 1. Monotonicity} : 
The distance between positional embeddings increases with larger time intervals.
\begin{align*}\label{monotonicity} 
\forall t_1, t_2, t_3 \in \mathbb{R}: ||t_1-t_2|| > ||t_1-t_3|| 
\\ \Leftrightarrow d(p(t_1), p(t_2)) > d(p(t_1), p(t_3)).
\end{align*}
This formally represents the requirement  for positional embeddings to preserve the sequential order of distances between positions. An embedding that adheres to monotonicity is commonly referred to as ordinal embedding. Monotonicity plays a crucial role in demonstrating that the embedding genuinely captures the positional information inherent in sequential data.

\noindent \textbf{Property 2. Translation invariance} : The distance between positional embeddings are translation invariant.
\begin{align*}
\forall t_1, t_2, \dots, t_n, l \in \mathbb{R}: d(p(t_1), p(t_1+l)) \\ = d(p(t_2), p(t_2+l)) = \dots = d(p(t_n), p(t_n+l)).
\end{align*}
Positional embeddings that consider the time gaps between observations should adhere to the property of translation invariance. This is especially important in time series tasks, where a long time series is cut into windows. The translation invariant property insures that the positional information is provided correctly in any way the window is created. \\

\noindent \textbf{Property 3. Symmetry} : The distance between positional embeddings are symmetric.
\begin{equation}\label{symmetry} 
\forall t_1, t_2 \in \mathbb{R}: d(p(t_1), p(t_2)) = d(p(t_2), p(t_1)).
\end{equation}
While symmetry seems natural for positional embeddings, relative positional embeddings do not meet this property and therefore is included. Symmetry is also closely related to inner product which also has symmetric property. The inner product is used many times in the self-attention of transformer. Therefore, the symmetric property of positional embedding is chosen.\\

We further include inductive and data-driven properties proposed from \cite{liu2020learning}.

\noindent \textbf{Property 4. Inductive} : The positional embedding is capable of accommodating sequences longer than those observed during training. For any maximum length $l$ of sequences at training time,
\begin{align*}\label{inductive} 
\{{\rm dom}(p_{train}): {\rm dom}(p_{train}) \in \mathbb{N},  {\rm dom}(p_{train}) \leq l \} \\ \rightarrow \{ {\rm dom}(p): {\rm dom}(p) \in \mathbb{N} \},
\end{align*}
where ${\rm dom}(f)$ denote the domain of function $f$. The inductive property allows the positional embedding to be created from shorter time windows, and handle longer windows necessary on test time. Furthermore, the inductive property shows that the postional embedding is not overfit to the training sequence length but is capable of handling variant sequence. \\ \\

\noindent \textbf{Property 5. Data-driven} : The positional embedding can be learned from the data. Depending on the periodicity of the time series, different positional embedding should be provided. The data-driven property ensures that the positional embedding is not fixed but can adaptively learn from the data. Sinusoidal embedding \cite{vaswani2017attention} is the representative of fixed embedding, and the feed-forward embedding used in BERT \cite{devlin2018bert} and many other pretrained large language models is a well-known example of data-driven learnable positional embedding. \\

Lastly, properties desired for irregular time series forecasting are included.

\noindent \textbf{Property 6. Irregularity-Adaptable} : The positional embedding can be defined on any irregularly spaced sequence.
\begin{equation}\label{irregular} 
{\rm dom}(p): {\rm dom}(p) \in \mathbb{R},  {\rm dom}(p) \leq l. 
\end{equation}
This is the most challenging property, since neural networks are created of regular learnable parameters. Therefore, the conventional positional embedding methods included the popularly used learnable embedding \cite{devlin2018bert} fails to provide positional embedding for irregular sequences. Sinusoidal positional embedding can be extended to handle irregular sequences, but it loses other properties of position while doing so. \\

\subsection{Visualizing sinusoidal and \algnamef.}
\begin{figure*}[h!]
    \centering
    \includegraphics[trim=0cm 13.5cm 0cm 0cm, width=\textwidth, clip]{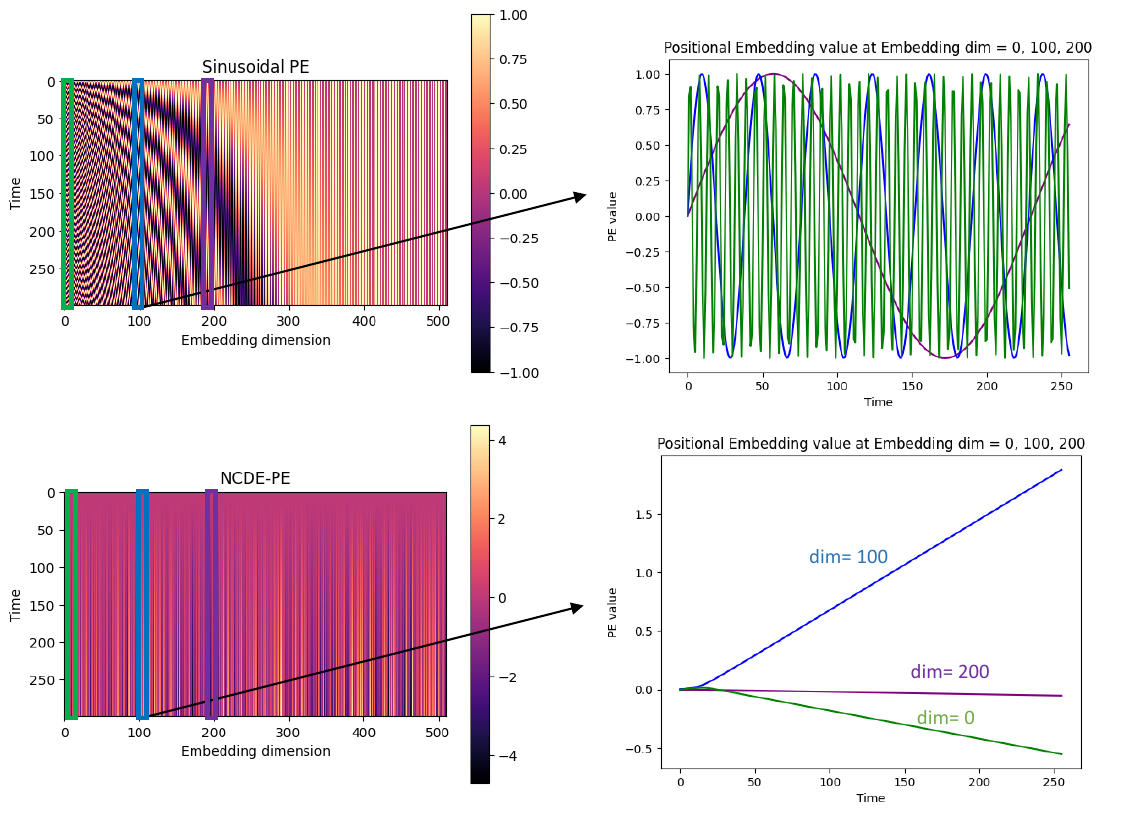}\\
    \caption{Sinusoidal positional embedding and \algnamef \ visualized.}
    \label{fig:PEvisual}
    \vspace*{-0.1cm}
\end{figure*}

%% file: tables/tab_ETTh1M.tex
\begin{table}[ht]
    \centering
    \caption{Forecasting results of ETT$\rm h_1$M based on Informer + RevIN.}
    \vspace*{0.2cm}
    \resizebox{\textwidth}{!}{%
    \begin{tabular}{@{} c|c | cc | cc | cc | cc | cc | cc | cc | cc | cc @{}}
    \toprule
    \multicolumn{2}{c|}{\textbf{Methods}} &
    \multicolumn{2}{c|}{\textbf{Uniform}} &
    \multicolumn{2}{c|}{\textbf{Sinusoidal}} &
    \multicolumn{2}{c|}{\textbf{Irr-sinusoidal}} &
    \multicolumn{2}{c|}{\textbf{Time Feature}} &
    \multicolumn{2}{c|}{\textbf{Simple}} &
    \multicolumn{2}{c|}{\textbf{Simple overlap}} &
    \multicolumn{2}{c|}{\textbf{\algnamef{}}}  &
    \multicolumn{2}{c|}{\textbf{Linear}} &
    \multicolumn{2}{c}{\textbf{Linear woB}} 
     \\\cline{1-20}
    Prediction Length & Missing Rate(\%) & MSE & MAE & MSE & MAE& MSE & MAE& MSE & MAE& MSE & MAE & MSE & MAE & MSE & MAE & MSE & MAE & MSE & MAE   \\

\midrule
\multirow{8}{*}{24}
& \multirow{2}{*}{0} 
& 0.519& 0.476& 0.470& 0.456& 0.502& 0.473& 0.808& 0.595& 0.459& 0.446& 0.479& 0.456& 0.402& 0.414& \textbf{0.394}& \textbf{0.410}& 0.411& 0.421\\
&& $(\pm0.029)$& $(\pm0.018)$& $(\pm0.033)$& $(\pm0.012)$& $(\pm0.020)$& $(\pm0.011)$& $(\pm0.036)$& $(\pm0.013)$& $(\pm0.033)$& $(\pm0.015)$& $(\pm0.051)$& $(\pm0.022)$& $(\pm0.012)$& $(\pm0.005)$& $(\pm0.005)$& $(\pm0.001)$& $(\pm0.009)$& $(\pm0.004)$\\& \multirow{2}{*}{20} 
& 0.699& 0.558& 0.681& 0.548& 0.697& 0.556& 0.746& 0.574& 0.682& 0.550& 0.700& 0.562& \textbf{0.633}& \textbf{0.517}& 0.636& 0.526& 0.634& 0.524\\
&& $(\pm0.029)$& $(\pm0.004)$& $(\pm0.028)$& $(\pm0.014)$& $(\pm0.013)$& $(\pm0.006)$& $(\pm0.018)$& $(\pm0.010)$& $(\pm0.004)$& $(\pm0.004)$& $(\pm0.025)$& $(\pm0.013)$& $(\pm0.029)$& $(\pm0.011)$& $(\pm0.017)$& $(\pm0.008)$& $(\pm0.006)$& $(\pm0.007)$\\& \multirow{2}{*}{40} 
& 0.824& 0.620& 0.839& 0.625& 0.815& 0.626& 0.796& 0.622& 0.837& 0.639& 0.849& 0.629& 0.764& 0.589& \textbf{0.741}& \textbf{0.587}& 0.749& 0.592\\
&& $(\pm0.014)$& $(\pm0.011)$& $(\pm0.042)$& $(\pm0.018)$& $(\pm0.041)$& $(\pm0.018)$& $(\pm0.032)$& $(\pm0.025)$& $(\pm0.039)$& $(\pm0.016)$& $(\pm0.029)$& $(\pm0.011)$& $(\pm0.013)$& $(\pm0.007)$& $(\pm0.004)$& $(\pm0.003)$& $(\pm0.008)$& $(\pm0.005)$\\& \multirow{2}{*}{60} 
& 0.809& 0.630& 0.879& 0.667& 0.887& 0.670& 0.874& 0.662& 0.847& 0.650& 0.909& 0.677& 0.737& 0.578& 0.724& 0.576& \textbf{0.718}& \textbf{0.560}\\
&& $(\pm0.016)$& $(\pm0.010)$& $(\pm0.033)$& $(\pm0.016)$& $(\pm0.034)$& $(\pm0.014)$& $(\pm0.031)$& $(\pm0.008)$& $(\pm0.041)$& $(\pm0.029)$& $(\pm0.050)$& $(\pm0.022)$& $(\pm0.013)$& $(\pm0.018)$& $(\pm0.006)$& $(\pm0.004)$& $(\pm0.003)$& $(\pm0.009)$\\
\midrule
\multirow{8}{*}{48}
& \multirow{2}{*}{0} 
& 0.562& 0.516& 0.613& 0.527& 0.551& 0.493& 0.870& 0.656& 0.647& 0.539& 0.643& 0.551& \textbf{0.492}& \textbf{0.469}& 0.509& 0.477& 0.555& 0.501\\
&& $(\pm0.039)$& $(\pm0.024)$& $(\pm0.014)$& $(\pm0.007)$& $(\pm0.044)$& $(\pm0.015)$& $(\pm0.072)$& $(\pm0.031)$& $(\pm0.072)$& $(\pm0.030)$& $(\pm0.061)$& $(\pm0.023)$& $(\pm0.020)$& $(\pm0.010)$& $(\pm0.013)$& $(\pm0.005)$& $(\pm0.004)$& $(\pm0.002)$\\& \multirow{2}{*}{20} 
& 0.815& 0.615& 0.837& 0.617& 0.850& 0.622& 0.811& 0.629& 0.921& 0.658& 0.808& 0.617& \textbf{0.723}& \textbf{0.566}& 0.733& 0.579& 0.734& 0.583\\
&& $(\pm0.071)$& $(\pm0.031)$& $(\pm0.068)$& $(\pm0.019)$& $(\pm0.064)$& $(\pm0.013)$& $(\pm0.014)$& $(\pm0.009)$& $(\pm0.055)$& $(\pm0.013)$& $(\pm0.029)$& $(\pm0.010)$& $(\pm0.028)$& $(\pm0.013)$& $(\pm0.016)$& $(\pm0.009)$& $(\pm0.013)$& $(\pm0.009)$\\& \multirow{2}{*}{40} 
& 0.889& 0.668& 0.869& 0.664& 0.905& 0.682& 0.916& 0.684& 0.880& 0.676& 1.036& 0.732& 0.765& 0.597& \textbf{0.720}& \textbf{0.565}& 0.723& 0.568\\
&& $(\pm0.060)$& $(\pm0.032)$& $(\pm0.033)$& $(\pm0.011)$& $(\pm0.040)$& $(\pm0.025)$& $(\pm0.102)$& $(\pm0.040)$& $(\pm0.025)$& $(\pm0.014)$& $(\pm0.073)$& $(\pm0.023)$& $(\pm0.009)$& $(\pm0.009)$& $(\pm0.005)$& $(\pm0.006)$& $(\pm0.013)$& $(\pm0.016)$\\& \multirow{2}{*}{60} 
& 0.831& 0.656& 0.914& 0.696& 0.903& 0.686& 0.864& 0.674& 0.956& 0.712& 0.846& 0.666& 0.721& 0.591& \textbf{0.692}& 0.559& 0.695& \textbf{0.557}\\
&& $(\pm0.033)$& $(\pm0.020)$& $(\pm0.033)$& $(\pm0.018)$& $(\pm0.058)$& $(\pm0.024)$& $(\pm0.064)$& $(\pm0.030)$& $(\pm0.071)$& $(\pm0.032)$& $(\pm0.043)$& $(\pm0.017)$& $(\pm0.009)$& $(\pm0.003)$& $(\pm0.007)$& $(\pm0.011)$& $(\pm0.017)$& $(\pm0.027)$\\
\midrule
\multirow{8}{*}{168}
& \multirow{2}{*}{0} 
& 0.727& 0.588& 0.733& 0.594& \textbf{0.669}& \textbf{0.562}& 0.933& 0.699& 0.738& 0.600& 0.773& 0.612& 0.845& 0.648& 0.748& 0.601& 0.780& 0.623\\
&& $(\pm0.026)$& $(\pm0.009)$& $(\pm0.038)$& $(\pm0.026)$& $(\pm0.019)$& $(\pm0.006)$& $(\pm0.015)$& $(\pm0.021)$& $(\pm0.066)$& $(\pm0.034)$& $(\pm0.074)$& $(\pm0.024)$& $(\pm0.236)$& $(\pm0.106)$& $(\pm0.004)$& $(\pm0.003)$& $(\pm0.010)$& $(\pm0.007)$\\& \multirow{2}{*}{20} 
& 1.023& 0.750& 1.110& 0.788& 1.058& 0.759& 1.073& 0.780& 1.060& 0.775& 1.069& 0.781& 0.922& 0.700& \textbf{0.729}& \textbf{0.588}& 0.751& 0.609\\
&& $(\pm0.007)$& $(\pm0.012)$& $(\pm0.061)$& $(\pm0.024)$& $(\pm0.081)$& $(\pm0.039)$& $(\pm0.070)$& $(\pm0.028)$& $(\pm0.161)$& $(\pm0.078)$& $(\pm0.056)$& $(\pm0.027)$& $(\pm0.104)$& $(\pm0.049)$& $(\pm0.022)$& $(\pm0.019)$& $(\pm0.016)$& $(\pm0.015)$\\& \multirow{2}{*}{40} 
& 1.101& 0.797& 1.141& 0.813& 1.165& 0.817& 1.030& 0.763& 1.168& 0.830& 1.256& 0.866& 0.828& 0.663& \textbf{0.718}& \textbf{0.589}& 0.725& 0.592\\
&& $(\pm0.024)$& $(\pm0.010)$& $(\pm0.093)$& $(\pm0.040)$& $(\pm0.100)$& $(\pm0.040)$& $(\pm0.078)$& $(\pm0.036)$& $(\pm0.030)$& $(\pm0.006)$& $(\pm0.016)$& $(\pm0.016)$& $(\pm0.048)$& $(\pm0.033)$& $(\pm0.011)$& $(\pm0.006)$& $(\pm0.015)$& $(\pm0.010)$\\& \multirow{2}{*}{60} 
& 0.937& 0.726& 0.954& 0.742& 0.938& 0.735& 0.937& 0.729& 1.055& 0.791& 1.043& 0.787& 0.794& 0.634& 0.692& 0.590& \textbf{0.691}& \textbf{0.587}\\
&& $(\pm0.054)$& $(\pm0.033)$& $(\pm0.012)$& $(\pm0.005)$& $(\pm0.032)$& $(\pm0.019)$& $(\pm0.089)$& $(\pm0.055)$& $(\pm0.087)$& $(\pm0.039)$& $(\pm0.033)$& $(\pm0.023)$& $(\pm0.091)$& $(\pm0.033)$& $(\pm0.017)$& $(\pm0.012)$& $(\pm0.008)$& $(\pm0.007)$\\
\midrule
\multirow{8}{*}{336}
& \multirow{2}{*}{0} 
& 0.859& 0.663& 0.840& 0.654& 0.996& 0.717& 0.988& 0.735& 1.105& 0.782& 1.173& 0.820& 0.815& 0.654& 0.795& 0.633& \textbf{0.769}& \textbf{0.614}\\
&& $(\pm0.094)$& $(\pm0.042)$& $(\pm0.050)$& $(\pm0.023)$& $(\pm0.101)$& $(\pm0.058)$& $(\pm0.047)$& $(\pm0.033)$& $(\pm0.014)$& $(\pm0.009)$& $(\pm0.017)$& $(\pm0.012)$& $(\pm0.029)$& $(\pm0.019)$& $(\pm0.032)$& $(\pm0.034)$& $(\pm0.029)$& $(\pm0.037)$\\& \multirow{2}{*}{20} 
& 1.005& 0.756& 1.096& 0.802& 1.086& 0.796& 0.988& 0.742& 1.142& 0.824& 1.125& 0.820& 0.865& 0.673& \textbf{0.718}& \textbf{0.597}& 0.734& 0.608\\
&& $(\pm0.061)$& $(\pm0.035)$& $(\pm0.071)$& $(\pm0.026)$& $(\pm0.084)$& $(\pm0.047)$& $(\pm0.059)$& $(\pm0.033)$& $(\pm0.060)$& $(\pm0.037)$& $(\pm0.053)$& $(\pm0.018)$& $(\pm0.213)$& $(\pm0.101)$& $(\pm0.034)$& $(\pm0.032)$& $(\pm0.047)$& $(\pm0.038)$\\& \multirow{2}{*}{40} 
& 1.140& 0.828& 1.214& 0.863& 1.185& 0.856& 1.129& 0.817& 1.102& 0.810& 1.264& 0.886& 0.780& 0.645& 0.719& 0.609& \textbf{0.719}& \textbf{0.608}\\
&& $(\pm0.060)$& $(\pm0.029)$& $(\pm0.033)$& $(\pm0.011)$& $(\pm0.116)$& $(\pm0.052)$& $(\pm0.087)$& $(\pm0.050)$& $(\pm0.028)$& $(\pm0.015)$& $(\pm0.074)$& $(\pm0.025)$& $(\pm0.072)$& $(\pm0.051)$& $(\pm0.005)$& $(\pm0.002)$& $(\pm0.004)$& $(\pm0.001)$\\& \multirow{2}{*}{60} 
& 1.016& 0.777& 1.031& 0.793& 1.017& 0.781& 1.072& 0.797& 0.975& 0.755& 1.075& 0.801& \textbf{0.799}& \textbf{0.654}& 0.817& 0.660& 0.861& 0.680\\
&& $(\pm0.088)$& $(\pm0.051)$& $(\pm0.051)$& $(\pm0.030)$& $(\pm0.080)$& $(\pm0.039)$& $(\pm0.055)$& $(\pm0.036)$& $(\pm0.120)$& $(\pm0.064)$& $(\pm0.109)$& $(\pm0.059)$& $(\pm0.028)$& $(\pm0.016)$& $(\pm0.069)$& $(\pm0.031)$& $(\pm0.080)$& $(\pm0.034)$\\

    \bottomrule
    \end{tabular}%
    }
    \label{tab:ETTH1M}
\end{table}

%% file: tables/tab_ETTh2M.tex
\begin{table}[ht]
    \centering
    \caption{Forecasting results of ETT$\rm h_2$M based on Informer + RevIN.}
    \vspace*{0.2cm}
    \resizebox{\textwidth}{!}{%
    \begin{tabular}{@{} c|c | cc | cc | cc | cc | cc | cc | cc | cc | cc @{}}
    \toprule
    \multicolumn{2}{c|}{\textbf{Methods}} &
    \multicolumn{2}{c|}{\textbf{Uniform}} &
    \multicolumn{2}{c|}{\textbf{Sinusoidal}} &
    \multicolumn{2}{c|}{\textbf{Irr-sinusoidal}} &
    \multicolumn{2}{c|}{\textbf{Time Feature}} &
    \multicolumn{2}{c|}{\textbf{Simple}} &
    \multicolumn{2}{c|}{\textbf{Simple overlap}} &
    \multicolumn{2}{c|}{\textbf{\algnamef{}}}  &
    \multicolumn{2}{c|}{\textbf{Linear}} &
    \multicolumn{2}{c}{\textbf{Linear woB}} 
     \\\cline{1-20}
    Prediction Length & Missing Rate(\%) & MSE & MAE & MSE & MAE& MSE & MAE& MSE & MAE& MSE & MAE & MSE & MAE & MSE & MAE & MSE & MAE & MSE & MAE   \\

    \midrule
\multirow{8}{*}{24}
& \multirow{2}{*}{0} 
& 0.237& 0.326& 0.245& 0.331& 0.243& 0.328& 0.279& 0.355& 0.256& 0.340& 0.257& 0.340& 0.313& 0.365& \textbf{0.197}& \textbf{0.288}& 0.198& 0.289\\
&& $(\pm0.009)$& $(\pm0.009)$& $(\pm0.010)$& $(\pm0.003)$& $(\pm0.004)$& $(\pm0.002)$& $(\pm0.020)$& $(\pm0.013)$& $(\pm0.011)$& $(\pm0.005)$& $(\pm0.007)$& $(\pm0.006)$& $(\pm0.163)$& $(\pm0.105)$& $(\pm0.005)$& $(\pm0.004)$& $(\pm0.003)$& $(\pm0.002)$\\& \multirow{2}{*}{20} 
& 0.276& 0.355& 0.293& 0.364& 0.311& 0.373& 0.316& 0.375& 0.282& 0.356& 0.268& 0.346& 0.242& 0.323& 0.230& 0.315& \textbf{0.229}& \textbf{0.314}\\
&& $(\pm0.019)$& $(\pm0.011)$& $(\pm0.007)$& $(\pm0.002)$& $(\pm0.020)$& $(\pm0.013)$& $(\pm0.014)$& $(\pm0.009)$& $(\pm0.006)$& $(\pm0.001)$& $(\pm0.009)$& $(\pm0.005)$& $(\pm0.013)$& $(\pm0.008)$& $(\pm0.001)$& $(\pm0.001)$& $(\pm0.002)$& $(\pm0.002)$\\& \multirow{2}{*}{40} 
& 0.335& 0.391& 0.322& 0.380& 0.338& 0.391& 0.352& 0.397& 0.340& 0.392& 0.314& 0.378& 0.294& 0.356& \textbf{0.276}& \textbf{0.348}& 0.278& 0.348\\
&& $(\pm0.029)$& $(\pm0.018)$& $(\pm0.004)$& $(\pm0.002)$& $(\pm0.025)$& $(\pm0.016)$& $(\pm0.012)$& $(\pm0.007)$& $(\pm0.012)$& $(\pm0.008)$& $(\pm0.015)$& $(\pm0.007)$& $(\pm0.017)$& $(\pm0.010)$& $(\pm0.005)$& $(\pm0.005)$& $(\pm0.008)$& $(\pm0.005)$\\& \multirow{2}{*}{60} 
& 0.418& 0.433& 0.389& 0.416& 0.389& 0.418& 0.439& 0.441& 0.435& 0.439& 0.416& 0.432& 0.366& 0.404& 0.358& 0.398& \textbf{0.349}& \textbf{0.394}\\
&& $(\pm0.036)$& $(\pm0.020)$& $(\pm0.022)$& $(\pm0.010)$& $(\pm0.024)$& $(\pm0.017)$& $(\pm0.018)$& $(\pm0.012)$& $(\pm0.017)$& $(\pm0.006)$& $(\pm0.029)$& $(\pm0.016)$& $(\pm0.025)$& $(\pm0.012)$& $(\pm0.024)$& $(\pm0.013)$& $(\pm0.017)$& $(\pm0.009)$\\
\midrule
\multirow{8}{*}{48}
& \multirow{2}{*}{0} 
& 0.368& 0.406& 0.335& 0.390& 0.356& 0.402& 0.378& 0.411& 0.361& 0.400& 0.362& 0.405& 0.335& 0.386& \textbf{0.310}& \textbf{0.367}& 0.325& 0.375\\
&& $(\pm0.028)$& $(\pm0.018)$& $(\pm0.018)$& $(\pm0.009)$& $(\pm0.021)$& $(\pm0.011)$& $(\pm0.030)$& $(\pm0.014)$& $(\pm0.023)$& $(\pm0.013)$& $(\pm0.021)$& $(\pm0.010)$& $(\pm0.015)$& $(\pm0.011)$& $(\pm0.027)$& $(\pm0.016)$& $(\pm0.024)$& $(\pm0.014)$\\& \multirow{2}{*}{20} 
& 0.378& 0.418& 0.434& 0.449& 0.406& 0.434& 0.448& 0.452& 0.415& 0.435& 0.446& 0.453& \textbf{0.338}& \textbf{0.389}& 0.362& 0.401& 0.353& 0.397\\
&& $(\pm0.011)$& $(\pm0.010)$& $(\pm0.020)$& $(\pm0.007)$& $(\pm0.006)$& $(\pm0.003)$& $(\pm0.004)$& $(\pm0.003)$& $(\pm0.032)$& $(\pm0.019)$& $(\pm0.025)$& $(\pm0.013)$& $(\pm0.030)$& $(\pm0.015)$& $(\pm0.036)$& $(\pm0.016)$& $(\pm0.012)$& $(\pm0.006)$\\& \multirow{2}{*}{40} 
& 0.492& 0.479& 0.491& 0.471& 0.494& 0.480& 0.555& 0.505& 0.493& 0.477& 0.482& 0.469& 0.492& 0.467& 0.448& 0.445& \textbf{0.397}& \textbf{0.423}\\
&& $(\pm0.019)$& $(\pm0.011)$& $(\pm0.069)$& $(\pm0.032)$& $(\pm0.022)$& $(\pm0.012)$& $(\pm0.008)$& $(\pm0.006)$& $(\pm0.040)$& $(\pm0.018)$& $(\pm0.025)$& $(\pm0.019)$& $(\pm0.148)$& $(\pm0.076)$& $(\pm0.002)$& $(\pm0.002)$& $(\pm0.013)$& $(\pm0.009)$\\& \multirow{2}{*}{60} 
& 0.485& 0.480& 0.485& 0.478& 0.456& 0.462& 0.496& 0.481& 0.480& 0.476& 0.484& 0.476& 0.450& 0.454& \textbf{0.426}& \textbf{0.446}& 0.450& 0.463\\
&& $(\pm0.018)$& $(\pm0.011)$& $(\pm0.024)$& $(\pm0.016)$& $(\pm0.019)$& $(\pm0.009)$& $(\pm0.022)$& $(\pm0.012)$& $(\pm0.026)$& $(\pm0.015)$& $(\pm0.007)$& $(\pm0.003)$& $(\pm0.033)$& $(\pm0.018)$& $(\pm0.021)$& $(\pm0.013)$& $(\pm0.073)$& $(\pm0.048)$\\
\midrule
\multirow{8}{*}{168}
& \multirow{2}{*}{0} 
& 0.602& 0.550& 0.629& 0.554& 0.731& 0.586& 0.592& 0.530& 0.698& 0.586& 0.721& 0.597& 0.559& 0.515& \textbf{0.442}& \textbf{0.465}& 0.495& 0.501\\
&& $(\pm0.086)$& $(\pm0.037)$& $(\pm0.067)$& $(\pm0.034)$& $(\pm0.214)$& $(\pm0.096)$& $(\pm0.046)$& $(\pm0.025)$& $(\pm0.026)$& $(\pm0.012)$& $(\pm0.073)$& $(\pm0.020)$& $(\pm0.069)$& $(\pm0.030)$& $(\pm0.055)$& $(\pm0.044)$& $(\pm0.123)$& $(\pm0.078)$\\& \multirow{2}{*}{20} 
& 0.499& 0.502& 0.568& 0.525& 0.498& 0.493& 0.545& 0.517& 0.543& 0.515& 0.583& 0.529& \textbf{0.452}& \textbf{0.467}& 0.515& 0.507& 0.522& 0.514\\
&& $(\pm0.042)$& $(\pm0.029)$& $(\pm0.096)$& $(\pm0.049)$& $(\pm0.015)$& $(\pm0.011)$& $(\pm0.034)$& $(\pm0.016)$& $(\pm0.015)$& $(\pm0.009)$& $(\pm0.071)$& $(\pm0.030)$& $(\pm0.034)$& $(\pm0.015)$& $(\pm0.002)$& $(\pm0.002)$& $(\pm0.024)$& $(\pm0.010)$\\& \multirow{2}{*}{40} 
& 0.542& 0.519& 0.554& 0.518& 0.523& 0.511& 0.504& 0.503& 0.528& 0.514& 0.628& 0.560& \textbf{0.467}& \textbf{0.486}& 0.508& 0.511& 0.523& 0.512\\
&& $(\pm0.077)$& $(\pm0.036)$& $(\pm0.066)$& $(\pm0.027)$& $(\pm0.048)$& $(\pm0.021)$& $(\pm0.033)$& $(\pm0.017)$& $(\pm0.050)$& $(\pm0.026)$& $(\pm0.041)$& $(\pm0.019)$& $(\pm0.024)$& $(\pm0.019)$& $(\pm0.018)$& $(\pm0.011)$& $(\pm0.030)$& $(\pm0.015)$\\& \multirow{2}{*}{60} 
& 0.621& 0.586& 0.562& 0.542& 0.566& 0.545& 0.564& 0.546& 0.589& 0.559& 0.569& 0.551& 0.546& 0.536& 0.542& 0.526& \textbf{0.540}& \textbf{0.525}\\
&& $(\pm0.032)$& $(\pm0.022)$& $(\pm0.011)$& $(\pm0.012)$& $(\pm0.035)$& $(\pm0.018)$& $(\pm0.007)$& $(\pm0.006)$& $(\pm0.018)$& $(\pm0.007)$& $(\pm0.010)$& $(\pm0.005)$& $(\pm0.020)$& $(\pm0.010)$& $(\pm0.015)$& $(\pm0.006)$& $(\pm0.016)$& $(\pm0.006)$\\
\midrule
\multirow{8}{*}{336}
& \multirow{2}{*}{0} 
& 0.672& 0.575& 0.703& 0.586& 0.678& 0.573& 0.606& 0.549& 0.744& 0.602& 0.785& 0.621& 0.518& 0.504& 0.442& 0.466& \textbf{0.416}& \textbf{0.456}\\
&& $(\pm0.017)$& $(\pm0.012)$& $(\pm0.065)$& $(\pm0.028)$& $(\pm0.013)$& $(\pm0.006)$& $(\pm0.063)$& $(\pm0.031)$& $(\pm0.047)$& $(\pm0.017)$& $(\pm0.061)$& $(\pm0.027)$& $(\pm0.041)$& $(\pm0.024)$& $(\pm0.043)$& $(\pm0.023)$& $(\pm0.016)$& $(\pm0.014)$\\& \multirow{2}{*}{20} 
& 0.581& 0.538& 0.598& 0.548& 0.614& 0.553& 0.571& 0.532& 0.828& 0.635& 0.737& 0.604& 0.556& 0.525& 0.520& 0.506& \textbf{0.469}& \textbf{0.483}\\
&& $(\pm0.044)$& $(\pm0.019)$& $(\pm0.009)$& $(\pm0.012)$& $(\pm0.076)$& $(\pm0.036)$& $(\pm0.069)$& $(\pm0.032)$& $(\pm0.164)$& $(\pm0.054)$& $(\pm0.062)$& $(\pm0.030)$& $(\pm0.057)$& $(\pm0.029)$& $(\pm0.072)$& $(\pm0.035)$& $(\pm0.044)$& $(\pm0.023)$\\& \multirow{2}{*}{40} 
& 0.556& 0.534& 0.651& 0.573& 0.627& 0.572& 0.633& 0.576& 0.859& 0.678& 0.821& 0.664& \textbf{0.508}& \textbf{0.503}& 0.569& 0.536& 0.551& 0.532\\
&& $(\pm0.049)$& $(\pm0.026)$& $(\pm0.081)$& $(\pm0.037)$& $(\pm0.014)$& $(\pm0.009)$& $(\pm0.059)$& $(\pm0.030)$& $(\pm0.148)$& $(\pm0.059)$& $(\pm0.030)$& $(\pm0.016)$& $(\pm0.065)$& $(\pm0.036)$& $(\pm0.022)$& $(\pm0.009)$& $(\pm0.051)$& $(\pm0.020)$\\& \multirow{2}{*}{60} 
& \textbf{0.684}& \textbf{0.602}& 0.721& 0.621& 0.740& 0.631& 0.710& 0.616& 0.933& 0.725& 0.923& 0.712& 0.837& 0.651& 0.719& 0.633& 0.720& 0.628\\
&& $(\pm0.043)$& $(\pm0.023)$& $(\pm0.048)$& $(\pm0.024)$& $(\pm0.031)$& $(\pm0.011)$& $(\pm0.049)$& $(\pm0.022)$& $(\pm0.027)$& $(\pm0.012)$& $(\pm0.029)$& $(\pm0.018)$& $(\pm0.159)$& $(\pm0.072)$& $(\pm0.049)$& $(\pm0.039)$& $(\pm0.089)$& $(\pm0.043)$\\

    \bottomrule
    \end{tabular}%
    }
    \label{tab:ETTH2M}
\end{table}

%% file: tables/tab_WTHM.tex
\begin{table}[ht]
    \centering
    \caption{Forecasting results of WTH based on Informer + RevIN.}
    \vspace*{0.2cm}
    \resizebox{\textwidth}{!}{%
    \begin{tabular}{@{} c|c | cc | cc | cc | cc | cc | cc | cc | cc | cc @{}}
    \toprule
    \multicolumn{2}{c|}{\textbf{Methods}} &
    \multicolumn{2}{c|}{\textbf{Uniform}} &
    \multicolumn{2}{c|}{\textbf{Sinusoidal}} &
    \multicolumn{2}{c|}{\textbf{Irr-sinusoidal}} &
    \multicolumn{2}{c|}{\textbf{Time Feature}} &
    \multicolumn{2}{c|}{\textbf{Simple}} &
    \multicolumn{2}{c|}{\textbf{Simple overlap}} &
    \multicolumn{2}{c|}{\textbf{\algnamef{}}}  &
    \multicolumn{2}{c|}{\textbf{Linear}} &
    \multicolumn{2}{c}{\textbf{Linear woB}} 
     \\\cline{1-20}
    Prediction Length & Missing Rate(\%) & MSE & MAE & MSE & MAE& MSE & MAE& MSE & MAE& MSE & MAE & MSE & MAE & MSE & MAE & MSE & MAE & MSE & MAE   \\

    \midrule
\multirow{8}{*}{24}
& \multirow{2}{*}{0} 
& 0.715& 0.612& 0.683& 0.595& 0.704& 0.606& 0.700& 0.606& 0.349& 0.389& 0.353& 0.395& 0.359& 0.394& 0.344& 0.382& \textbf{0.336}& \textbf{0.377}\\
&& $(\pm0.031)$& $(\pm0.013)$& $(\pm0.013)$& $(\pm0.005)$& $(\pm0.021)$& $(\pm0.012)$& $(\pm0.010)$& $(\pm0.011)$& $(\pm0.008)$& $(\pm0.006)$& $(\pm0.001)$& $(\pm0.002)$& $(\pm0.010)$& $(\pm0.008)$& $(\pm0.002)$& $(\pm0.001)$& $(\pm0.004)$& $(\pm0.003)$\\& \multirow{2}{*}{20} 
& 0.733& 0.615& 0.691& 0.604& 0.730& 0.628& 0.745& 0.624& 0.434& 0.462& 0.425& 0.448& 0.441& 0.468& 0.416& 0.444& \textbf{0.414}& \textbf{0.443}\\
&& $(\pm0.043)$& $(\pm0.008)$& $(\pm0.006)$& $(\pm0.012)$& $(\pm0.027)$& $(\pm0.015)$& $(\pm0.047)$& $(\pm0.019)$& $(\pm0.019)$& $(\pm0.012)$& $(\pm0.004)$& $(\pm0.006)$& $(\pm0.027)$& $(\pm0.020)$& $(\pm0.012)$& $(\pm0.008)$& $(\pm0.018)$& $(\pm0.012)$\\& \multirow{2}{*}{40} 
& 0.759& 0.635& 0.744& 0.637& 0.759& 0.640& 0.748& 0.625& 0.481& 0.496& 0.458& 0.479& 0.481& 0.489& \textbf{0.450}& \textbf{0.474}& 0.464& 0.481\\
&& $(\pm0.054)$& $(\pm0.018)$& $(\pm0.020)$& $(\pm0.006)$& $(\pm0.017)$& $(\pm0.005)$& $(\pm0.010)$& $(\pm0.007)$& $(\pm0.018)$& $(\pm0.012)$& $(\pm0.005)$& $(\pm0.006)$& $(\pm0.015)$& $(\pm0.010)$& $(\pm0.005)$& $(\pm0.003)$& $(\pm0.012)$& $(\pm0.013)$\\& \multirow{2}{*}{60} 
& 0.771& 0.647& 0.717& 0.624& 0.751& 0.643& 0.722& 0.638& 0.555& 0.543& \textbf{0.522}& 0.528& 0.548& 0.529& 0.532& 0.520& 0.526& \textbf{0.517}\\
&& $(\pm0.009)$& $(\pm0.009)$& $(\pm0.009)$& $(\pm0.001)$& $(\pm0.043)$& $(\pm0.027)$& $(\pm0.027)$& $(\pm0.016)$& $(\pm0.010)$& $(\pm0.007)$& $(\pm0.016)$& $(\pm0.007)$& $(\pm0.031)$& $(\pm0.023)$& $(\pm0.049)$& $(\pm0.030)$& $(\pm0.055)$& $(\pm0.034)$\\
\midrule
\multirow{8}{*}{48}
& \multirow{2}{*}{0} 
& 0.700& 0.598& 0.686& 0.592& 0.706& 0.604& 0.698& 0.597& 0.432& 0.442& 0.426& 0.444& 0.420& \textbf{0.432}& \textbf{0.419}& 0.435& 0.419& 0.436\\
&& $(\pm0.022)$& $(\pm0.011)$& $(\pm0.024)$& $(\pm0.009)$& $(\pm0.011)$& $(\pm0.006)$& $(\pm0.039)$& $(\pm0.016)$& $(\pm0.006)$& $(\pm0.003)$& $(\pm0.007)$& $(\pm0.003)$& $(\pm0.011)$& $(\pm0.008)$& $(\pm0.003)$& $(\pm0.001)$& $(\pm0.003)$& $(\pm0.003)$\\& \multirow{2}{*}{20} 
& 0.736& 0.617& 0.724& 0.611& 0.720& 0.612& 0.701& 0.605& 0.513& 0.503& 0.496& 0.491& 0.534& 0.510& \textbf{0.480}& \textbf{0.482}& 0.483& 0.484\\
&& $(\pm0.009)$& $(\pm0.004)$& $(\pm0.007)$& $(\pm0.005)$& $(\pm0.010)$& $(\pm0.003)$& $(\pm0.008)$& $(\pm0.005)$& $(\pm0.010)$& $(\pm0.006)$& $(\pm0.003)$& $(\pm0.002)$& $(\pm0.030)$& $(\pm0.012)$& $(\pm0.005)$& $(\pm0.003)$& $(\pm0.004)$& $(\pm0.003)$\\& \multirow{2}{*}{40} 
& 0.734& 0.622& 0.724& 0.622& 0.731& 0.623& 0.755& 0.633& 0.548& 0.531& 0.546& 0.528& 0.543& 0.527& 0.512& 0.508& \textbf{0.510}& \textbf{0.504}\\
&& $(\pm0.025)$& $(\pm0.009)$& $(\pm0.005)$& $(\pm0.006)$& $(\pm0.006)$& $(\pm0.007)$& $(\pm0.008)$& $(\pm0.004)$& $(\pm0.008)$& $(\pm0.007)$& $(\pm0.010)$& $(\pm0.005)$& $(\pm0.006)$& $(\pm0.007)$& $(\pm0.001)$& $(\pm0.006)$& $(\pm0.003)$& $(\pm0.003)$\\& \multirow{2}{*}{60} 
& 0.777& 0.651& 0.797& 0.652& 0.747& 0.638& 0.771& 0.644& 0.621& 0.581& 0.590& 0.562& 0.597& 0.559& 0.548& 0.530& \textbf{0.544}& \textbf{0.528}\\
&& $(\pm0.039)$& $(\pm0.018)$& $(\pm0.007)$& $(\pm0.007)$& $(\pm0.017)$& $(\pm0.006)$& $(\pm0.034)$& $(\pm0.015)$& $(\pm0.003)$& $(\pm0.002)$& $(\pm0.009)$& $(\pm0.001)$& $(\pm0.035)$& $(\pm0.018)$& $(\pm0.001)$& $(\pm0.002)$& $(\pm0.002)$& $(\pm0.001)$\\
\midrule
\multirow{8}{*}{168}
& \multirow{2}{*}{0} 
& 0.781& 0.654& 0.777& 0.656& 0.789& 0.666& 0.767& 0.654& 0.667& 0.606& 0.667& 0.599& 0.682& 0.600& 0.693& 0.610& \textbf{0.650}& \textbf{0.589}\\
&& $(\pm0.028)$& $(\pm0.016)$& $(\pm0.014)$& $(\pm0.009)$& $(\pm0.049)$& $(\pm0.023)$& $(\pm0.025)$& $(\pm0.006)$& $(\pm0.036)$& $(\pm0.022)$& $(\pm0.032)$& $(\pm0.013)$& $(\pm0.016)$& $(\pm0.006)$& $(\pm0.019)$& $(\pm0.020)$& $(\pm0.007)$& $(\pm0.008)$\\& \multirow{2}{*}{20} 
& 0.786& 0.660& 0.780& 0.660& 0.782& 0.658& 0.782& 0.658& 0.682& 0.612& 0.669& 0.608& 0.658& 0.597& 0.672& 0.604& \textbf{0.602}& \textbf{0.569}\\
&& $(\pm0.038)$& $(\pm0.018)$& $(\pm0.019)$& $(\pm0.004)$& $(\pm0.018)$& $(\pm0.003)$& $(\pm0.013)$& $(\pm0.009)$& $(\pm0.023)$& $(\pm0.015)$& $(\pm0.017)$& $(\pm0.013)$& $(\pm0.030)$& $(\pm0.014)$& $(\pm0.048)$& $(\pm0.024)$& $(\pm0.015)$& $(\pm0.010)$\\& \multirow{2}{*}{40} 
& 0.818& 0.677& 0.838& 0.685& 0.841& 0.689& 0.816& 0.675& 0.660& 0.610& 0.667& 0.611& 0.669& 0.603& 0.625& 0.580& \textbf{0.611}& \textbf{0.576}\\
&& $(\pm0.023)$& $(\pm0.012)$& $(\pm0.014)$& $(\pm0.007)$& $(\pm0.034)$& $(\pm0.009)$& $(\pm0.039)$& $(\pm0.019)$& $(\pm0.020)$& $(\pm0.011)$& $(\pm0.027)$& $(\pm0.013)$& $(\pm0.028)$& $(\pm0.011)$& $(\pm0.031)$& $(\pm0.023)$& $(\pm0.011)$& $(\pm0.007)$\\& \multirow{2}{*}{60} 
& 0.879& 0.709& 0.869& 0.704& 0.896& 0.719& 0.893& 0.714& 0.653& 0.600& 0.649& 0.598& 0.738& 0.641& 0.640& 0.601& \textbf{0.633}& \textbf{0.595}\\
&& $(\pm0.023)$& $(\pm0.017)$& $(\pm0.037)$& $(\pm0.016)$& $(\pm0.039)$& $(\pm0.014)$& $(\pm0.038)$& $(\pm0.011)$& $(\pm0.023)$& $(\pm0.012)$& $(\pm0.007)$& $(\pm0.005)$& $(\pm0.167)$& $(\pm0.078)$& $(\pm0.020)$& $(\pm0.015)$& $(\pm0.017)$& $(\pm0.013)$\\
\midrule
\multirow{8}{*}{336}
& \multirow{2}{*}{0} 
& 0.843& 0.692& 0.839& 0.694& 0.850& 0.698& 0.848& 0.697& \textbf{0.650}& \textbf{0.603}& 0.674& 0.607& 0.729& 0.639& 0.729& 0.644& 0.786& 0.688\\
&& $(\pm0.010)$& $(\pm0.007)$& $(\pm0.009)$& $(\pm0.004)$& $(\pm0.007)$& $(\pm0.006)$& $(\pm0.009)$& $(\pm0.002)$& $(\pm0.010)$& $(\pm0.006)$& $(\pm0.031)$& $(\pm0.010)$& $(\pm0.058)$& $(\pm0.026)$& $(\pm0.034)$& $(\pm0.017)$& $(\pm0.065)$& $(\pm0.040)$\\& \multirow{2}{*}{20} 
& 0.891& 0.720& 0.873& 0.710& 0.904& 0.726& 0.902& 0.729& 0.674& 0.603& 0.682& 0.607& 0.666& \textbf{0.602}& 0.677& 0.611& \textbf{0.658}& 0.603\\
&& $(\pm0.018)$& $(\pm0.008)$& $(\pm0.014)$& $(\pm0.008)$& $(\pm0.016)$& $(\pm0.007)$& $(\pm0.021)$& $(\pm0.009)$& $(\pm0.046)$& $(\pm0.019)$& $(\pm0.009)$& $(\pm0.005)$& $(\pm0.019)$& $(\pm0.008)$& $(\pm0.016)$& $(\pm0.012)$& $(\pm0.019)$& $(\pm0.011)$\\& \multirow{2}{*}{40} 
& 0.984& 0.778& 0.986& 0.771& 0.982& 0.771& 1.006& 0.778& 0.660& 0.607& \textbf{0.645}& \textbf{0.600}& 0.820& 0.686& 0.708& 0.638& 0.646& 0.608\\
&& $(\pm0.010)$& $(\pm0.006)$& $(\pm0.026)$& $(\pm0.016)$& $(\pm0.008)$& $(\pm0.003)$& $(\pm0.008)$& $(\pm0.003)$& $(\pm0.045)$& $(\pm0.020)$& $(\pm0.021)$& $(\pm0.013)$& $(\pm0.130)$& $(\pm0.078)$& $(\pm0.134)$& $(\pm0.079)$& $(\pm0.026)$& $(\pm0.018)$\\& \multirow{2}{*}{60} 
& 1.263& 0.902& 1.229& 0.889& 1.231& 0.891& 1.233& 0.897& 0.676& 0.620& \textbf{0.642}& \textbf{0.604}& 0.957& 0.768& 1.898& 1.102& 0.800& 0.683\\
&& $(\pm0.037)$& $(\pm0.020)$& $(\pm0.011)$& $(\pm0.007)$& $(\pm0.026)$& $(\pm0.012)$& $(\pm0.020)$& $(\pm0.004)$& $(\pm0.026)$& $(\pm0.010)$& $(\pm0.019)$& $(\pm0.013)$& $(\pm0.233)$& $(\pm0.108)$& $(\pm0.170)$& $(\pm0.054)$& $(\pm0.291)$& $(\pm0.148)$\\

    \bottomrule
    \end{tabular}%
    }
    \label{tab:WTHM}
\end{table}

%% file: tables/tab_fed_ETTh1M.tex
\begin{table}[ht]
    \centering
    \caption{Forecasting results of ETT$\rm h_1$M based on FEDformer + RevIN.}
    \vspace*{0.2cm}
    \resizebox{\textwidth}{!}{%
    \begin{tabular}{@{} c|c | cc | cc | cc | cc | cc @{}}
    \toprule
    \multicolumn{2}{c|}{\textbf{Methods}} &
    \multicolumn{2}{c|}{\textbf{Irr-sinusoidal}} &
    \multicolumn{2}{c|}{\textbf{Sinusoidal}} &
    \multicolumn{2}{c|}{\textbf{Time Feature}} &
    \multicolumn{2}{c|}{\textbf{Linear}} &
    \multicolumn{2}{c}{\textbf{Linear woB}} 
     \\\cline{1-12}
    Prediction Length & Missing Rate(\%) & MSE & MAE & MSE & MAE& MSE & MAE& MSE & MAE& MSE & MAE   \\

\midrule
\multirow{8}{*}{96}
& \multirow{2}{*}{0} 
& 0.394& \textbf{0.405}& \textbf{0.393}& 0.406& 0.571& 0.514& 0.397& 0.410& 0.397& 0.412\\
&& $(\pm0.002)$& $(\pm0.001)$& $(\pm0.002)$& $(\pm0.001)$& $(\pm0.001)$& $(\pm0.002)$& $(\pm0.003)$& $(\pm0.002)$& $(\pm0.008)$& $(\pm0.002)$\\& \multirow{2}{*}{20} 
& 0.649& 0.534& \textbf{0.641}& \textbf{0.528}& 0.650& 0.547& 0.776& 0.585& 0.737& 0.568\\
&& $(\pm0.003)$& $(\pm0.002)$& $(\pm0.003)$& $(\pm0.001)$& $(\pm0.005)$& $(\pm0.002)$& $(\pm0.009)$& $(\pm0.004)$& $(\pm0.013)$& $(\pm0.005)$\\& \multirow{2}{*}{40} 
& 0.714& \textbf{0.559}& 0.715& 0.559& \textbf{0.698}& 0.564& 1.114& 0.686& 1.036& 0.668\\
&& $(\pm0.005)$& $(\pm0.004)$& $(\pm0.002)$& $(\pm0.003)$& $(\pm0.050)$& $(\pm0.023)$& $(\pm0.064)$& $(\pm0.018)$& $(\pm0.031)$& $(\pm0.005)$\\& \multirow{2}{*}{60} 
& 0.716& 0.570& 0.697& 0.566& \textbf{0.618}& \textbf{0.530}& 1.055& 0.679& 1.020& 0.676\\
&& $(\pm0.013)$& $(\pm0.005)$& $(\pm0.002)$& $(\pm0.001)$& $(\pm0.014)$& $(\pm0.006)$& $(\pm0.118)$& $(\pm0.037)$& $(\pm0.051)$& $(\pm0.014)$\\
\midrule
\multirow{8}{*}{192}
& \multirow{2}{*}{0} 
& 0.466& \textbf{0.441}& 0.465& 0.441& 0.666& 0.552& 0.461& 0.443& \textbf{0.461}& 0.446\\
&& $(\pm0.004)$& $(\pm0.001)$& $(\pm0.002)$& $(\pm0.001)$& $(\pm0.010)$& $(\pm0.004)$& $(\pm0.003)$& $(\pm0.000)$& $(\pm0.004)$& $(\pm0.003)$\\& \multirow{2}{*}{20} 
& 0.708& 0.567& \textbf{0.696}& \textbf{0.558}& 0.752& 0.580& 1.002& 0.669& 1.097& 0.696\\
&& $(\pm0.010)$& $(\pm0.003)$& $(\pm0.013)$& $(\pm0.009)$& $(\pm0.024)$& $(\pm0.010)$& $(\pm0.089)$& $(\pm0.028)$& $(\pm0.102)$& $(\pm0.028)$\\& \multirow{2}{*}{40} 
& 0.747& 0.587& \textbf{0.733}& \textbf{0.579}& 0.775& 0.588& 1.242& 0.725& 1.318& 0.749\\
&& $(\pm0.003)$& $(\pm0.004)$& $(\pm0.005)$& $(\pm0.004)$& $(\pm0.023)$& $(\pm0.007)$& $(\pm0.095)$& $(\pm0.031)$& $(\pm0.112)$& $(\pm0.024)$\\& \multirow{2}{*}{60} 
& 0.715& 0.584& 0.693& 0.579& \textbf{0.651}& \textbf{0.555}& 0.987& 0.658& 1.122& 0.698\\
&& $(\pm0.013)$& $(\pm0.009)$& $(\pm0.006)$& $(\pm0.004)$& $(\pm0.012)$& $(\pm0.003)$& $(\pm0.357)$& $(\pm0.093)$& $(\pm0.317)$& $(\pm0.091)$\\
\midrule
\multirow{8}{*}{336}
& \multirow{2}{*}{0} 
& \textbf{0.497}& \textbf{0.463}& 0.502& 0.466& 0.778& 0.588& 0.500& 0.465& 0.505& 0.465\\
&& $(\pm0.011)$& $(\pm0.007)$& $(\pm0.008)$& $(\pm0.003)$& $(\pm0.025)$& $(\pm0.013)$& $(\pm0.005)$& $(\pm0.008)$& $(\pm0.008)$& $(\pm0.003)$\\& \multirow{2}{*}{20} 
& 0.740& 0.585& \textbf{0.727}& \textbf{0.574}& 0.809& 0.601& 1.249& 0.739& 1.411& 0.782\\
&& $(\pm0.004)$& $(\pm0.003)$& $(\pm0.006)$& $(\pm0.004)$& $(\pm0.001)$& $(\pm0.005)$& $(\pm0.055)$& $(\pm0.013)$& $(\pm0.023)$& $(\pm0.009)$\\& \multirow{2}{*}{40} 
& 0.757& 0.608& \textbf{0.732}& \textbf{0.596}& 0.825& 0.623& 1.338& 0.759& 1.138& 0.702\\
&& $(\pm0.001)$& $(\pm0.002)$& $(\pm0.009)$& $(\pm0.005)$& $(\pm0.073)$& $(\pm0.028)$& $(\pm0.303)$& $(\pm0.078)$& $(\pm0.329)$& $(\pm0.092)$\\& \multirow{2}{*}{60} 
& 0.751& 0.615& 0.757& 0.627& \textbf{0.709}& \textbf{0.590}& 0.745& 0.613& 1.085& 0.689\\
&& $(\pm0.009)$& $(\pm0.006)$& $(\pm0.011)$& $(\pm0.006)$& $(\pm0.007)$& $(\pm0.002)$& $(\pm0.017)$& $(\pm0.006)$& $(\pm0.328)$& $(\pm0.091)$\\
\midrule
\multirow{8}{*}{720}
& \multirow{2}{*}{0} 
& \textbf{0.528}& \textbf{0.486}& 0.528& 0.486& 0.838& 0.629& 0.550& 0.499& 0.562& 0.508\\
&& $(\pm0.020)$& $(\pm0.008)$& $(\pm0.020)$& $(\pm0.008)$& $(\pm0.035)$& $(\pm0.010)$& $(\pm0.023)$& $(\pm0.010)$& $(\pm0.023)$& $(\pm0.007)$\\& \multirow{2}{*}{20} 
& 0.807& 0.637& \textbf{0.802}& \textbf{0.634}& 0.890& 0.654& 1.412& 0.789& 1.600& 0.834\\
&& $(\pm0.027)$& $(\pm0.013)$& $(\pm0.017)$& $(\pm0.011)$& $(\pm0.036)$& $(\pm0.014)$& $(\pm0.092)$& $(\pm0.027)$& $(\pm0.046)$& $(\pm0.010)$\\& \multirow{2}{*}{40} 
& 0.975& 0.702& \textbf{0.922}& \textbf{0.691}& 1.167& 0.777& 1.253& 0.755& 1.715& 0.862\\
&& $(\pm0.017)$& $(\pm0.007)$& $(\pm0.027)$& $(\pm0.010)$& $(\pm0.089)$& $(\pm0.024)$& $(\pm0.329)$& $(\pm0.086)$& $(\pm0.561)$& $(\pm0.134)$\\& \multirow{2}{*}{60} 
& 1.103& 0.775& 1.210& 0.812& \textbf{0.960}& 0.733& 1.189& 0.793& 0.984& \textbf{0.723}\\
&& $(\pm0.045)$& $(\pm0.014)$& $(\pm0.044)$& $(\pm0.012)$& $(\pm0.078)$& $(\pm0.026)$& $(\pm0.095)$& $(\pm0.036)$& $(\pm0.027)$& $(\pm0.010)$\\

    \bottomrule
    \end{tabular}%
    }
    \label{tab:FED}
\end{table}

%% file: example_paper.bbl
\begin{thebibliography}{30}
\providecommand{\natexlab}[1]{#1}
\providecommand{\url}[1]{\texttt{#1}}
\expandafter\ifx\csname urlstyle\endcsname\relax
  \providecommand{\doi}[1]{doi: #1}\else
  \providecommand{\doi}{doi: \begingroup \urlstyle{rm}\Url}\fi

\bibitem[Brown et~al.(2020)Brown, Mann, Ryder, Subbiah, Kaplan, Dhariwal, Neelakantan, Shyam, Sastry, Askell, et~al.]{brown2020language}
Brown, T., Mann, B., Ryder, N., Subbiah, M., Kaplan, J.~D., Dhariwal, P., Neelakantan, A., Shyam, P., Sastry, G., Askell, A., et~al.
\newblock Language models are few-shot learners.
\newblock \emph{Advances in neural information processing systems}, 33:\penalty0 1877--1901, 2020.

\bibitem[Che et~al.(2018)Che, Purushotham, Cho, Sontag, and Liu]{che2018recurrent}
Che, Z., Purushotham, S., Cho, K., Sontag, D., and Liu, Y.
\newblock Recurrent neural networks for multivariate time series with missing values.
\newblock \emph{Scientific reports}, 8\penalty0 (1):\penalty0 6085, 2018.

\bibitem[Chen et~al.(2018)Chen, Rubanova, Bettencourt, and Duvenaud]{chen2018neural}
Chen, R.~T., Rubanova, Y., Bettencourt, J., and Duvenaud, D.~K.
\newblock Neural ordinary differential equations.
\newblock \emph{Advances in neural information processing systems}, 31, 2018.

\bibitem[Choi et~al.(2016)Choi, Bahadori, Schuetz, Stewart, and Sun]{choi2016doctor}
Choi, E., Bahadori, M.~T., Schuetz, A., Stewart, W.~F., and Sun, J.
\newblock Doctor ai: Predicting clinical events via recurrent neural networks.
\newblock In \emph{Machine learning for healthcare conference}, pp.\  301--318. PMLR, 2016.

\bibitem[Dal~Pozzolo et~al.(2015)Dal~Pozzolo, Caelen, Johnson, and Bontempi]{dal2015calibrating}
Dal~Pozzolo, A., Caelen, O., Johnson, R.~A., and Bontempi, G.
\newblock Calibrating probability with undersampling for unbalanced classification.
\newblock In \emph{2015 IEEE symposium series on computational intelligence}, pp.\  159--166. IEEE, 2015.

\bibitem[Devlin et~al.(2018)Devlin, Chang, Lee, and Toutanova]{devlin2018bert}
Devlin, J., Chang, M.-W., Lee, K., and Toutanova, K.
\newblock Bert: Pre-training of deep bidirectional transformers for language understanding.
\newblock \emph{arXiv preprint arXiv:1810.04805}, 2018.

\bibitem[He et~al.(2016)He, Zhang, Ren, and Sun]{he2016identity}
He, K., Zhang, X., Ren, S., and Sun, J.
\newblock Identity mappings in deep residual networks.
\newblock In \emph{European conference on computer vision}, pp.\  630--645. Springer, 2016.

\bibitem[Kidger(2022)]{kidger2022neural}
Kidger, P.
\newblock On neural differential equations.
\newblock \emph{arXiv preprint arXiv:2202.02435}, 2022.

\bibitem[Kidger et~al.(2020)Kidger, Morrill, Foster, and Lyons]{kidger2020neural}
Kidger, P., Morrill, J., Foster, J., and Lyons, T.
\newblock Neural controlled differential equations for irregular time series.
\newblock \emph{Advances in Neural Information Processing Systems}, 33:\penalty0 6696--6707, 2020.

\bibitem[Kim et~al.(2021)Kim, Kim, Tae, Park, Choi, and Choo]{kim2021reversible}
Kim, T., Kim, J., Tae, Y., Park, C., Choi, J.-H., and Choo, J.
\newblock Reversible instance normalization for accurate time-series forecasting against distribution shift.
\newblock In \emph{International Conference on Learning Representations}, 2021.

\bibitem[Li et~al.(2019)Li, Jin, Xuan, Zhou, Chen, Wang, and Yan]{li2019enhancing}
Li, S., Jin, X., Xuan, Y., Zhou, X., Chen, W., Wang, Y.-X., and Yan, X.
\newblock Enhancing the locality and breaking the memory bottleneck of transformer on time series forecasting.
\newblock \emph{Advances in neural information processing systems}, 32, 2019.

\bibitem[Lipton et~al.(2016)Lipton, Kale, and Wetzel]{lipton2016directly}
Lipton, Z.~C., Kale, D., and Wetzel, R.
\newblock Directly modeling missing data in sequences with rnns: Improved classification of clinical time series.
\newblock In \emph{Machine learning for healthcare conference}, pp.\  253--270. PMLR, 2016.

\bibitem[Liu et~al.(2021)Liu, Yu, Liao, Li, Lin, Liu, and Dustdar]{liu2021pyraformer}
Liu, S., Yu, H., Liao, C., Li, J., Lin, W., Liu, A.~X., and Dustdar, S.
\newblock Pyraformer: Low-complexity pyramidal attention for long-range time series modeling and forecasting.
\newblock In \emph{Proceedings of the International Conference on Learning Representations}, 2021.

\bibitem[Liu et~al.(2020)Liu, Yu, Dhillon, and Hsieh]{liu2020learning}
Liu, X., Yu, H.-F., Dhillon, I., and Hsieh, C.-J.
\newblock Learning to encode position for transformer with continuous dynamical model.
\newblock In \emph{International conference on machine learning}, pp.\  6327--6335. PMLR, 2020.

\bibitem[Menne et~al.(2015)Menne, Williams~Jr, and Vose]{menne2015united}
Menne, M.~J., Williams~Jr, C., and Vose, R.~S.
\newblock United states historical climatology network daily temperature, precipitation, and snow data.
\newblock \emph{Carbon Dioxide Information Analysis Center, Oak Ridge National Laboratory, Oak Ridge, Tennessee}, 2015.

\bibitem[Nie et~al.(2023)Nie, Nguyen, Sinthong, and Kalagnanam]{nie2023a}
Nie, Y., Nguyen, N.~H., Sinthong, P., and Kalagnanam, J.
\newblock A time series is worth 64 words: Long-term forecasting with transformers.
\newblock In \emph{The Eleventh International Conference on Learning Representations}, 2023.
\newblock URL \url{https://openreview.net/forum?id=Jbdc0vTOcol}.

\bibitem[Radford(2018)]{radford2018improving}
Radford, A.
\newblock Improving language understanding by generative pre-training.
\newblock \emph{arXiv preprint}, 2018.

\bibitem[Rubanova et~al.(2019)Rubanova, Chen, and Duvenaud]{rubanova2019latent}
Rubanova, Y., Chen, R.~T., and Duvenaud, D.~K.
\newblock Latent ordinary differential equations for irregularly-sampled time series.
\newblock \emph{Advances in neural information processing systems}, 32, 2019.

\bibitem[Salinas et~al.(2020)Salinas, Flunkert, Gasthaus, and Januschowski]{salinas2020deepar}
Salinas, D., Flunkert, V., Gasthaus, J., and Januschowski, T.
\newblock Deepar: Probabilistic forecasting with autoregressive recurrent networks.
\newblock \emph{International Journal of Forecasting}, 36\penalty0 (3):\penalty0 1181--1191, 2020.

\bibitem[Shaw et~al.(2018)Shaw, Uszkoreit, and Vaswani]{shaw2018self}
Shaw, P., Uszkoreit, J., and Vaswani, A.
\newblock Self-attention with relative position representations.
\newblock \emph{arXiv preprint arXiv:1803.02155}, 2018.

\bibitem[Shukla \& Marlin(2021)Shukla and Marlin]{shukla2021multi}
Shukla, S.~N. and Marlin, B.~M.
\newblock Multi-time attention networks for irregularly sampled time series.
\newblock \emph{arXiv preprint arXiv:2101.10318}, 2021.

\bibitem[Silva et~al.(2012)Silva, Moody, Scott, Celi, and Mark]{silva2012predicting}
Silva, I., Moody, G., Scott, D.~J., Celi, L.~A., and Mark, R.~G.
\newblock Predicting in-hospital mortality of icu patients: The physionet/computing in cardiology challenge 2012.
\newblock In \emph{2012 Computing in Cardiology}, pp.\  245--248. IEEE, 2012.

\bibitem[Trindade(2015)]{misc_electricityloaddiagrams20112014_321}
Trindade, A.
\newblock {ElectricityLoadDiagrams20112014}.
\newblock UCI Machine Learning Repository, 2015.
\newblock {DOI}: https://doi.org/10.24432/C58C86.

\bibitem[Vaswani et~al.(2017)Vaswani, Shazeer, Parmar, Uszkoreit, Jones, Gomez, Kaiser, and Polosukhin]{vaswani2017attention}
Vaswani, A., Shazeer, N., Parmar, N., Uszkoreit, J., Jones, L., Gomez, A.~N., Kaiser, {\L}., and Polosukhin, I.
\newblock Attention is all you need.
\newblock In \emph{Proceedings of Advances in Neural Information Processing Systems}, volume~30, 2017.

\bibitem[Wang et~al.(2020)Wang, Shang, Lioma, Jiang, Yang, Liu, and Simonsen]{wang2020position}
Wang, B., Shang, L., Lioma, C., Jiang, X., Yang, H., Liu, Q., and Simonsen, J.~G.
\newblock On position embeddings in bert.
\newblock In \emph{International Conference on Learning Representations}, 2020.

\bibitem[Wen et~al.(2022)Wen, Zhou, Zhang, Chen, Ma, Yan, and Sun]{wen2022transformers}
Wen, Q., Zhou, T., Zhang, C., Chen, W., Ma, Z., Yan, J., and Sun, L.
\newblock Transformers in time series: A survey.
\newblock \emph{arXiv preprint arXiv:2202.07125}, 2022.

\bibitem[Wu et~al.(2021)Wu, Xu, Wang, and Long]{wu2021autoformer}
Wu, H., Xu, J., Wang, J., and Long, M.
\newblock Autoformer: Decomposition transformers with auto-correlation for long-term series forecasting.
\newblock \emph{Advances in Neural Information Processing Systems}, 34:\penalty0 22419--22430, 2021.

\bibitem[Zeng et~al.(2023)Zeng, Chen, Zhang, and Xu]{zeng2023transformers}
Zeng, A., Chen, M., Zhang, L., and Xu, Q.
\newblock Are transformers effective for time series forecasting?
\newblock In \emph{Proceedings of the Thirty-Seventh AAAI Conference on Artificial Intelligence and Thirty-Fifth Conference on Innovative Applications of Artificial Intelligence and Thirteenth Symposium on Educational Advances in Artificial Intelligence}, pp.\  11121--11128, 2023.

\bibitem[Zhou et~al.(2021)Zhou, Zhang, Peng, Zhang, Li, Xiong, and Zhang]{zhou2021informer}
Zhou, H., Zhang, S., Peng, J., Zhang, S., Li, J., Xiong, H., and Zhang, W.
\newblock Informer: Beyond efficient transformer for long sequence time-series forecasting.
\newblock In \emph{35th AAAI Conference on Artificial Intelligence, AAAI 2021}, pp.\  11106--11115. Association for the Advancement of Artificial Intelligence, 2021.

\bibitem[Zhou et~al.(2022)Zhou, Ma, Wen, Wang, Sun, and Jin]{zhou2022fedformer}
Zhou, T., Ma, Z., Wen, Q., Wang, X., Sun, L., and Jin, R.
\newblock {FEDFormer}: Frequency enhanced decomposed transformer for long-term series forecasting.
\newblock In \emph{Proceedings of the International Conference on Machine Learning}, pp.\  27268--27286, 2022.

\end{thebibliography}
